\documentclass[journal]{IEEEtran}
\usepackage{cite}
\usepackage{tabularx, graphicx,
  amsmath,amsfonts,siunitx,pifont,amssymb,subcaption}
\usepackage{xcolor, bm, float, multirow, multicol, courier, mathtools, url}
\usepackage[linesnumbered,ruled,vlined]{algorithm2e} \pdfminorversion=4
 \usepackage{enumitem} \usepackage{graphicx}
\usepackage{balance}
\newcommand{\etal}{\textit{et al}.~}

\DeclareMathOperator*{\argmin}{argmin}
\DeclareMathOperator*{\argmax}{argmax}
\newcommand{\abs}[1]{\lvert#1\rvert}

\usepackage{siunitx}
\usepackage{arydshln}
\usepackage{pdfpages}
\usepackage{cite}
\usepackage{algpseudocode}
\usepackage{float}
\usepackage{colortbl}
\usepackage{dblfloatfix}    
\usepackage{mathrsfs}
\usepackage{color}
\usepackage[colorinlistoftodos]{todonotes}

\def\secref#1{Sec.~\ref{#1}}
\def\figref#1{Fig.~\ref{#1}}
\def\tabref#1{Tab.~\ref{#1}}
\def\eqref#1{Eq.~(\ref{#1})}

\ifCLASSINFOpdf
\else
\fi
\begin{document}
%
\title{AEROS: AdaptivE RObust least-Squares for Graph-Based SLAM}


\author{\IEEEauthorblockN{Milad Ramezani\,$^{1}$,~\IEEEmembership{Member,~IEEE},
Matias Mattamala\,$^{2}$,~\IEEEmembership{Student Member,~IEEE}, and
Maurice Fallon\,,$^{2}$,~\IEEEmembership{Member,~IEEE}}

\IEEEauthorblockA{$^{1}$ Robotics and Autonomous Systems Group, DATA61, CSIRO, Brisbane,
Australia\\$^{2}$ Dynamic Robot Systems, Oxford Robotics Institute, Department of Engineering
Science, University of
Oxford, Oxford, UK}


\thanks{
Corresponding author: M. Ramezani (email: milad@robots.ox.ac.uk).}}

%




\maketitle

\begin{abstract}
In robot localisation and mapping, outliers are unavoidable when loop-closure measurements are
taken into account.
A single false-positive loop-closure can have a very negative
impact on SLAM problems causing an inferior trajectory to be produced or even 
for the optimisation to fail entirely.
To address this issue, popular existing approaches define a hard switch for 
each loop-closure constraint.
This paper presents AEROS, a novel approach to adaptively 
solve a robust least-squares minimisation problem by adding just a single extra 
latent parameter. It can be used in the back-end component of
the SLAM problem to enable generalised robust cost minimisation by 
simultaneously estimating the continuous latent parameter along with the set of 
sensor poses in a single joint optimisation. This leads to a very closely curve
fitting on the distribution 
of the residuals, thereby reducing the effect of outliers. Additionally, we
formulate the robust optimisation problem using standard Gaussian factors so 
that it can be solved by direct application of popular incremental 
estimation approaches such as iSAM.
Experimental results on publicly available synthetic datasets and real LiDAR-SLAM datasets collected
from the 2D and 3D LiDAR systems show the competitiveness of our approach with the state-of-the-art
techniques and its superiority on real world scenarios.
\end{abstract}

\begin{IEEEkeywords}
Robust cost function, outlier resilience, back-end optimisation, factor graph, least squares
minimisation, SLAM, perception.
\end{IEEEkeywords}


%
\IEEEpeerreviewmaketitle

\section{Introduction}
%
%
%
%

\IEEEPARstart{S}{tate-of-the-art} SLAM systems commonly use sparse pose graphs which are a
topological structure made up of nodes representing the poses of the robot/sensor at certain times, and edges
representing constraints between those poses. The constraints are the probabilistic
expression of either the relative translation between consecutive poses (from an odometry system) or a loop-closure
between non-consecutive poses when an explored area is revisited. These constraints are typically
determined using
a SLAM front-end that converts the raw sensor measurements into constraints added to the
pose graph sequentially.
The back-end of the graph-based SLAM system estimates the optimal pose 
trajectory by solving a non-linear least squares minimisation problem.
Packages such as GTSAM~\cite{dellaert2012factor},
$g^2o$~\cite{kummerle2011g} and the Ceres solver~\cite{agarwal2012ceres} have been developed
for this back-end optimisation.

Erroneous constraints can occur, particularly when determining loop-closures. False-positive
loop-closure constraints can cause the optimiser to converge to an incorrect solution resulting in a
twisted trajectory and a corrupted map. This, in turn, affects other downstream components such as
path planning, collision avoidance and terrain mapping. Hence, the detection of outliers and the
mitigation of their negative impact is an important problem.

One way to do this is to sanitise the constraints in the front-end --- with high thresholds needed to
establish a loop closure. Outlier rejection can be strengthened by accumulating
a belief propagated over several constraints~\cite{cummins2008fab,
cadena2012robust}, using statistical
hypotheses~\cite{olson2009recognizing,olson2012progress} or by detecting environmentally degenerate
conditions~\cite{zhang2016degeneracy, nobili2018predicting,
ramezani2020online}. There are limits to this approach, with careful and
conservative hand-tuning needed to avoid rejecting true loop closures 
which ought to be detected.

While the front-end aims to remove ill-conditioned constraints during the data association, 
it is typically a separate processing unit and revisiting a constraint involves reprocessing raw sensor measurements.
The alternative has been to add robustness to the back-end of the SLAM system.
Back-end outlier rejection is a more abstract problem which is coupled to the pose-graph
optimiser --- which is the specific problem at hand.  


One approach has been to reduce the effect of outliers 
with adjustments to the least squares minimiser so that they have minimal effect during 
optimisation. 
Promising works such as 
Switchable Constraints (SC)~\cite{sunderhauf2012towards}, Dynamic Covariance Scaling
(DCS)~\cite{agarwal2013robust}, Max-Mixture (MM)~\cite{olson2013mix} and Graduated
Non-Convexity (GNC) non-minimal solvers~\cite{yang2020graduated} have been 
proposed to
make parameter estimation robust to outliers, each of which have 
limitations that we discuss further
in~\secref{related-work}. As an example, the SC approach requires an additional variable for every
loop-closure added to the pose-graph; it also cannot effectively deactivate 
outliers when the
outlier-to-inlier ratio exceeds a threshold~\cite{sunderhauf2012robust}. The GNC-based algorithm
relies on a specific M-estimator~\cite{zhang1997}, meaning that algorithm is not general for
different scenarios.

In this paper we present
AdaptivE RObust least-Squares -- AEROS. Our primary contribution is an adaptive robust cost
function which can be used to represent the entire set of M-estimators by
estimating a hyper-parameter in a joint
optimisation along with the pose parameters. By leveraging the Black-Rangarajan
duality~\cite{black1996unification} and benefiting from the generalised robust cost function
proposed by Barron~\cite{Barron_2019_CVPR}, we demonstrate that the robust cost function can be
rewritten
in the form of a standard least squares minimisation.
We aim to jointly estimate the hyper-parameter with the sensor poses. For this porpuse, it is not possible to directly use the
Barron’s robust kernel in a least squares minimisation (as proven in \cite{chebrolu2020adaptive}).
As a result we take advantage of the Black-Rangarajan duality to convert the optimization problem
into an Iteratively Reweighted Least Squares (IRLS) problem instead.

We test our approach using standard publicly available datasets and real 3D data
collected using a LiDAR handheld device in both structured
and unstructured environments. The
result from our algorithm is then evaluated against ground truth trajectories and 
compared with other outlier-rejection approaches, namely the SC, DCS, GNC algorithms and one of
the M-estimators Geman-McClure (GM). We
show that
our approach can handle a large outlier-to-inlier ratio which is
likely to
occur when, for instance, visual information is used for loop-closure
detection~\cite{cadena2016past}.

The remainder of the paper is organised as follows.~\secref{related-work} 
discusses previous
approaches that
deal with outliers either in the front-end or in the back-end. ~\secref{dynamic-graph} presents our
proposed method for adaptive dynamic pose-graph SLAM --- beginning with a brief introduction to 
Switchable Constraints in~\secref{sc} and concluding with our proposed adaptive factors.
In~\secref{sec:experiments}, we evaluate the performance of our 
proposed approach using publicly available synthetic datasets as well as with real data generated by
our LiDAR-SLAM
system, before presenting conclusions in~\secref{conclusion}.

\section{Related Work}
\label{related-work}
Many works have been developed to deal with false positive when closing loops in SLAM. These approaches can be
divided into two groups depending on whether they sanitise the constraints in the front end or add optimisation robustness in
the back end. This section overviews the body of related work according to that split.

\subsection{Robust Front-End}
Appearance-based place recognition algorithms have been developed to propose loop closures in Visual SLAM
(V-SLAM). Visual bag-of-words, originally proposed in~\cite{sivic2003video}, match a compound summary of
the image to propose image pairs which ought to correspond to the same place.
However, relying only on visual information is challenging due to 
perceptual aliasing, requiring additional verification steps to confirm a loop 
closure candidate. 

To robustify the appearance-based place recognition against outliers, Cummins 
and Newman~\cite{cummins2008fab,cummins2011appearance} proposed a probabilistic 
framework for appearance-based navigation
called FAB-MAP. Robust data association is achieved by combining information 
about the current pose as well as probabilistic models of the places. 
Cadena~\etal~\cite{cadena2012robust} improved the robustness of their appearance-based loop closure system
by performing a variety of similarity and temporal consistency checks followed 
by a Conditional Random Field (CRFs)~\cite{lafferty2001} verification.
Similar approaches have been used for data association of 
visual~\cite{ramos2009learning} or LiDAR 
measurements~\cite{burgard2008crf}. 

Olson~\cite{olson2009recognizing} proposed a place recognition method which 
uses spectrally clustered local matches to achieve global consistency.
Local uniqueness guaranteed that the
loop closure is unique to a particular neighbourhood, while the
global sufficiency verifies the unambiguity of the local match within the 
positional uncertainty of the robot. 
Later Olson~\etal~\cite{olson2012progress} proposed a simple loop-closure 
verification method using a temporary pose graph. It included all the candidate 
loops and performed a Chi-squared test (aka $\chi^2$ test) before adding them 
to the final pose graph to be optimised.

There are also methods that reject unreliable loop closures by analysing the structure of the
environment. Zhang~\etal~\cite{zhang2016degeneracy} estimated odometry 
constraints by determining and
separating degenerate dimensions in the state space. In this way, the 
optimisation was constrained to only the well-conditioned dimensions. 
Nobili~\etal~\cite{nobili2018predicting} predicted the alignment risk of two 
partly-overlapping 
laser scans, given the initial poses of the scans from which they are captured. 
While they introduced a metric to detect degenerate dimensions, this approach 
was not implemented for loop-closure verification. Since the approach 
was not suitable for real-time operation due to the exhaustive search needed to 
find overlapped surfaces, later Ramezani~\etal~\cite{ramezani2020online} 
accelerated this method by using a kd-tree.

As a conclusion, front-end robustness requires application dependent tuning and 
configuration, affecting its generalisation to different scenarios with 
different sensor configurations. This makes robust back-end methods more 
appealing, since they provide a general method to deal with outliers.

\subsection{Robust Back-End}
In contrast to robust front-end methods, in these methods the responsibility 
for determining the validity of loop closures lies on the SLAM back-end, i.e. 
the optimiser.

A method which has gained in popularity is Switchable Constraints (SC) proposed by
S\"{u}nderhauf and Protzel~\cite{sunderhauf2012towards}. As described in~\secref{sc}, for every
loop closure edge added to the pose graph, a switch variable is added to the optimisation. The
switch variables are estimated in a joint optimisation with the poses. The objective function of
the associated loop closures will be down-weighted if they are erroneous so as to mitigate the
impact of outliers. This approach, however, increases computation complexity since each loop
closure constraint is associated with this secondary variable.
To avoid these additional switch variables, Agarwal~\etal~\cite{agarwal2013robust} suggested
Dynamic Covariance Scaling (DCS) whose central idea is the same as SC. Instead 
of adding an additional switch variable for every loop closure, a closed-form 
solution is used to select a switch factor for each loop closure.
Since the formulation is proportional
to the original error of the loop closure constraint, the information
matrix of the loop closure constraints is dynamically reweighted without 
requiring extra computation.

Olson and Agarwal~\cite{olson2013mix} suggested using multi-modal distributions 
in the maximum likelihood instead of unimodal Gaussians. With this in mind, 
they replaced the sum-mixture
Gaussian with a max-mixture Gaussian and used this approach to detect uncertain loop closures. By
defining two components for each loop closure, i.e. the loop closure being computed from the
front-end and a null hypothesis which represents the case in which the loop closure is false, the
authors allow the optimiser to select a more plausible state. However, this approach does not
guarantee convergence to the global optimum.

Latif~\etal~\cite{latif2013robust} introduced an algorithm, called Realizing, Reversing and
Recovering (RRR), which clustered the pose graph edges, i.e. constraints consistent with one
another, using the place recognition constraints.
They conducted two $\chi^2$ tests to check the consistency of clusters with each other
and with the odometry links.
A difference between the RRR algorithm and the two previously mentioned
approaches~\cite{agarwal2013robust,olson2013mix} is that RRR makes a binary
decision and the suspicious loop closure will be omitted if a statistical criteria is not met. In
contrast, the other approaches keep the outlier in the optimisation, but with a minimal weight.

M-estimators~\cite{zhang1997} are the standard technique for robust optimisation in robotics.
They aim to remove the effect of outliers when fitting a model by
replacing the standard square error with a robust function that has
the same behaviour in a basin close to zero error but have a lower penalising effect
outside the basin. Some M-estimators which are commonly used in robotics include
Huber~\cite{huber1992robust}, Cauchy~\cite{black1996robust} and L1-L2~\cite{zhang1997}.

There are two problems with M-estimators: their parameters need to be manually tuned and one
M-estimator cannot generalise to all the problems, i.e. the selection of an M-estimator depends on
the problem and the data.
Resolving these problems was the motivation
for Agamennoni~\etal\cite{agamennoni2015self}. They developed an approach of self-tuning
M-estimators in which the tuning parameter of the M-estimator was considered as a variable to
be itself estimated in the optimisation problem within a two-step Expectation-Maximisation (EM)
procedure in an iterative fashion. Although this approach can tune the parameters of the M-estimator
automatically, the optimal choice of M-estimator family was still an expert decision. The
authors concluded that Student-\textit{t} M-estimator was a much better fit to the residuals than
the other
estimators they tested, albeit with testing on a limited number of datasets.

Another approach, which used the EM algorithm, was the work of Chebrolu~\etal~\cite{chebrolu2020adaptive}.
Instead of estimating the optimal tuning parameter of a specific M-estimator, the authors estimated
the shape parameter of a general robust function, proposed by Barron~\cite{Barron_2019_CVPR}, to
fit the best curve to the probability distribution of the residuals. 
Nonetheless, their algorithm
produces a sub-optimal solution because the hyperparameters (either the tuning 
parameter or the shape parameter) and the main parameters (robot poses) are not 
jointly estimated.

Recently, Yang~\etal~\cite{yang2020graduated} proposed an approach to outlier rejection that used
an M-estimator and converted it to a series of outlier processes, described
in~\secref{black-rang}, by leveraging the Black-Rangarajan duality~\cite{black1996unification}. The
authors used Graduated Non-Convexity (GNC) to achieve a robust estimation without any
initialisation. The key idea of GNC is to start with a convex problem and replace the robust
function with a surrogate function governed by a control parameter and to gradually recover the
original cost function. Nevertheless, this approach needs to update the 
hyperparameters, in this
case the weights, in an update step similar to the EM algorithm. In addition, 
the choice of M-estimator was a user decision.

In the same context, Lee~\etal~\cite{lee2013robust} showed that the corresponding weights for
the loop closure constraints can be chosen using a Cauchy function so that the optimisation becomes
convex. They also reweighted the original residuals in an EM procedure.
Carlone~\etal~\cite{carlone2014selecting} used an L1 proxy to relax a non-convex L0
optimisation to
find the largest subset of coherent and observable measurements. However, as noted by the authors,
applying convex relaxation does not guarantee the maximum-cardinality of the subset.

Finally, there have also been efforts that breakdown the distinction between front-end and back-end
to jointly attack the problem. For example, Wu and 
Beltrame~\cite{wu2020cluster} presented a
cluster-based approach that
leverages the spatial clustering technique in~\cite{latif2013robust} to group loop closures
topologically close to one another. They checked the
local consistency of each cluster in the front-end with a $\chi^2$ test, and then they applied a
Cluster-based Penalty Scaling (CPS) along with switchable constraints to achieve global consistency
in the back-end. However, their approach requires more computation time than either SC or DCS. This
complexity is due to the need to examine each cluster in the front-end and adding the clustered-based
penalty to the back-end.

\section{Dynamic Pose-Graph SLAM}
\label{dynamic-graph}

A sparse pose graph typically is made up of odometry and loop-closure (LC) factors. Adding these factors
to the graph, we seek a Maximum A Posterior (MAP) estimate of the set of poses $\mathcal{X}$. In a
LiDAR-SLAM system, the poses correspond to the sensor poses from which the laser scans were
collected. To produce a MAP estimate
over the variables $\mathcal{X}_i$, the product of all the factors must be 
maximised:
\begin{equation}
\label{eq:map_max}
\mathcal{X}^{\mathcal{MAP}} = \ \argmax_{\mathcal{X}} \ \prod_i
\underbrace{\phi_i(\mathcal{X}_{i-1},\mathcal{X}_{i})}_\text{Odometry
Factor} \prod_j
\underbrace{\phi_j(\mathcal{X}_{p_j},\mathcal{X}_{q_j})}_\text{LC Factor},
\vspace{-0.5em}
\end{equation}
where the odometry factors are defined between consecutive poses $\mathcal{X}_i$ and
$\mathcal{X}_{i-1}$, and $\mathcal{X}_{p_j}$ and $\mathcal{X}_{q_j}$ are the tail and head poses of
the \textit{jth}
loop closure.

Assuming that error is Gaussian distributed, and using the Negative Log Likelihood (NLL), the optimal variable configuration
in~\eqref{eq:map_max} can be determined by minimising a nonlinear least squares 
problem:
\begin{equation}
\label{eq:map_min}
\begin{split}
\mathcal{X}^{\mathcal{MAP}} = \ \argmin_{\mathcal{X}} \sum_{i}
\| f_i(\mathcal{X}_i,\mathcal{X}_{i-1})-y_i\|^2_{\bm{\Sigma}_i} \\ +
\sum_{j}
\|f_j(\mathcal{X}_{p_j},\mathcal{X}_{q_j})-y_j\|^2_{\bm{\Lambda}_j},
\end{split}
\end{equation}
where $y$ and $f$ denote the measurements and their corresponding non-linear models. Matrices
$\bm{\Sigma}$ and $\bm{\Lambda}$ are the covariance matrices corresponding to the odometry factors
and loop-closures, respectively.

\subsection{Least Squares with Switch Variables}
\label{sc}
An efficient approach when dealing with false-positive loop-closures in the back-end is the approach proposed
by S{\"u}nderhauf and Protzel~\cite{sunderhauf2012towards} in which, for each loop-closure added to
the factor graph, a switch variable is added to reweight the cost function:

\begin{equation}
\label{eq:map_min_sc}
\begin{split}
\mathcal{\{X, S\}}^{\mathcal{MAP}} = \ \argmin_{\mathcal{X, S}} \sum_{i}
\| f_i(\mathcal{X}_i,\mathcal{X}_{i-1})-y_i\|^2_{\bm{\Sigma}_i} \\ +
\sum_{j}\bigg(
\underbrace{w(s_j)\|f_j(\mathcal{X}_{p_j},\mathcal{X}_{q_j})-y_j\|^2_{\bm{\Lambda}_j}}_\text{
Switched LC
Factor} +
\underbrace{\|s_j-\lambda\|^2_{\sigma^2_j}}_\text{Switch Prior}\bigg),
\end{split}
\end{equation}
in which, the summations can be interpreted as three independent residuals of the odometry,
switched loop-closure and switch prior factors. $\mathcal{S}$ is a vector of switch
variables, $s_j$, being jointly estimated along with $\mathcal{X}$. $w(s_j) \in 
[0,1]$ denotes the
weighting
function which can be either a simple linear function, e.g. $w(s_j)=s_j$ or a non-linear function
such as sigmoid, i.e. $w(s_j)=\frac{1}{(1+\exp(-s_j))}$~\cite{sunderhauf2012robust}. The scalar
$\lambda$ represents an initial guess of the switch variable around which $s_j$ can be adjusted
within a standard deviation, $\sigma_j$, such that the entire range of $w$ is covered.

As noted in~\cite{sunderhauf2012towards}, all loop-closures are initially treated as being true-positive,
i.e. as inliers, thus the initial guess $\lambda$ must be selected such that the weight $w$
tends to 1. In the case of using a linear function, e.g. $w(s_j)=s_j$, $\lambda$ should be set to
1 with $\sigma_j=1$. If a sigmoid function is used, the initial value of $s_j$ is selected as 10
with $\sigma_j=20$. This setting guarantees that the range $[-5,5]$, where $w$ varies from 0 to 1,
is covered~\cite{sunderhauf2012towards}.

When using the switchable loop-closure constraints, the switch prior
is required because only this term behaves as a penalty to prevent the minimisation
from driving switch variables of the inliers to zero. Otherwise, it is the natural 
behaviour of the minimisation to disable all the loop-closures.

The drawback of switchable constraints is that for each loop-closure constraint, a
switch variable must be added to the joint optimisation. As a result, as the number of switch
variables increases, in addition to the complexity of the SLAM problem increasing, some outliers
are likely not to be down-weighed, resulting in global distortion~\cite{sunderhauf2012robust}
(see~\figref{fig:3D-sphere}).

\begin{figure}[t]
\centering
 \begin{minipage}{.5\textwidth}
 \centering
   \includegraphics[width=1.0\linewidth]{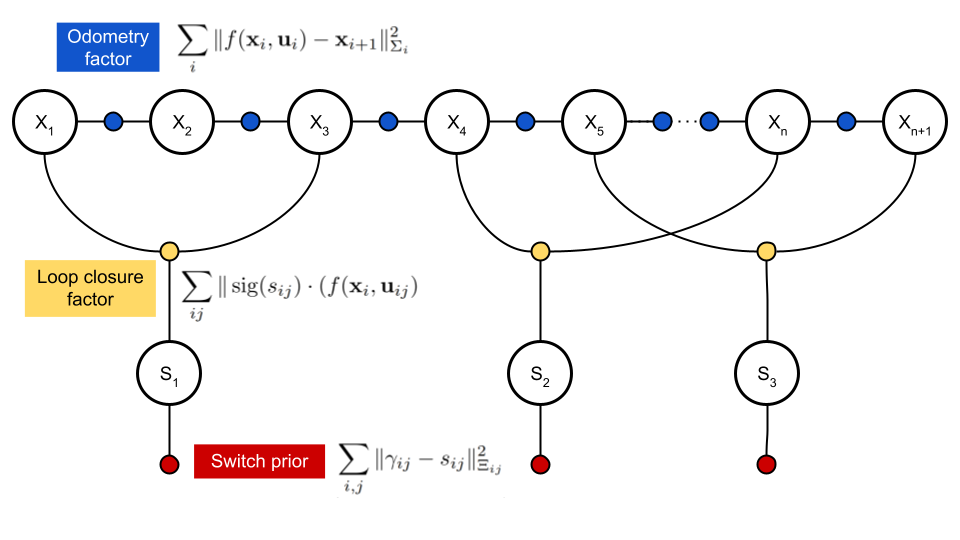}\\
   \vspace{1mm}
  \end{minipage}
  \begin{minipage}{.5\textwidth}
  \centering
  \vspace{5mm}
   \includegraphics[width=1.0\linewidth]{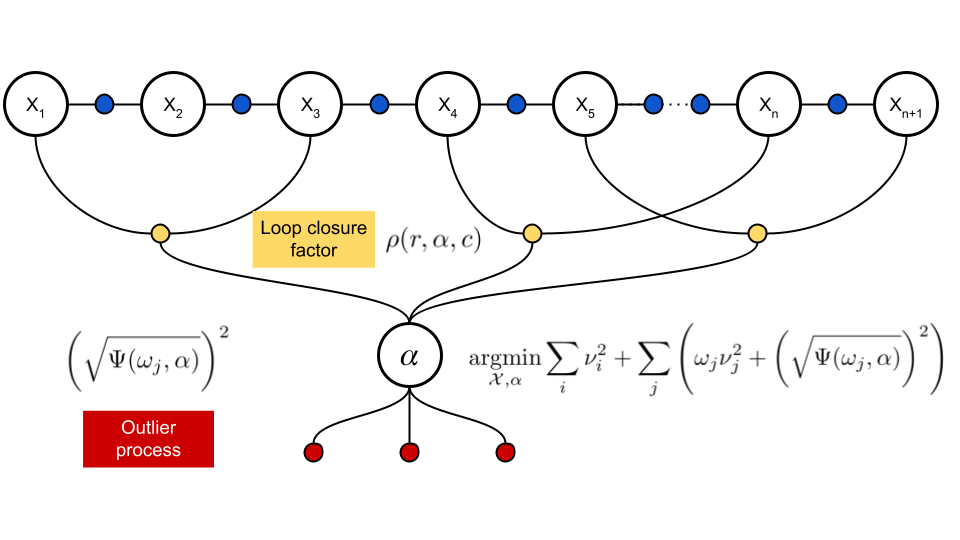}
  \end{minipage}
     \caption{\small{Representation of loop closing factors and their additional components added
to the graph by the SC method (top) and our proposal (bottom). As seen, in the 
SC structure, each switchable loop-closure is accompanied with 
a switch variable
($\text{s}_i$) and a switch prior. In contrast, in the structure of AEROS, all the adaptive loop closures are connected to the latent parameter ($\alpha$) and for every loop closure
constraint an outlier process is defined. }}
\label{fig:sc-ad}
\vspace{-2mm}
\end{figure}

\subsection{Least Squares with M-estimators}
\label{robust-cost}

Another widely used technique to reduce the impact of false-positive loop 
closures on the optimisation is
M-estimation~\cite{zhang1997}. Using robust M-estimation, \eqref{eq:map_min} can be rewritten by
replacing the quadratic cost function in the loop closure term by a
robust function of residuals:

\begin{equation}
\label{eq:map_robust}
\begin{split}
\mathcal{X}^{\mathcal{MAP}} = \ \argmin_{\mathcal{X}} \sum_{i}
\| f_i(\mathcal{X}_i,\mathcal{X}_{i-1})-y_i\|^2_{\bm{\Sigma}_i} \\ +
\sum_{j}
\rho\big(\|f_j(\mathcal{X}_{p_j},\mathcal{X}_{q_j})-y_j\|_{\bm{\Lambda}^\frac{1}{2}_j}\big)
\end{split}
\end{equation}
here, $\rho$ is a symmetric, positive-definite function which is convex around zero and
has smaller gradient than the quadratic function when far from the minimum, 
therefore it is less forgiving about outliers.

Throughout the paper, we assume that odometry constraints are reliable and only loop-closure
constraints are susceptible to
outliers, thus we define a robust cost function for that term. Defining
$\nu_j=\|f_j(\mathcal{X}_{p_j},\mathcal{X}_{q_j})-y_j\|_{\bm{\Lambda}^\frac{1}{2}_j}$,
~\eqref{eq:map_robust} yields:

\begin{equation}
\label{eq:m_estimators}
\argmin_{\mathcal{X}}{\sum_{j}
\rho(\nu_j)}
\end{equation}

The M-estimator of $\mathcal{X}$ based
upon the robust cost function $\rho(\nu_j)$ is the solution of the following equation:

\begin{equation}
\label{eq:influential}
\sum_{j}\psi(\nu_j)\frac{\partial \nu_j}{\partial\mathcal{X}_i}=0,\ \text{for}\ i=1,\dots,m,
\end{equation}
where, $\psi(\nu)=\frac{d\rho(\nu)}{d\nu}$, known as the influence function, is 
the derivative of the
robust function $\rho$ with respect to the residual $\nu$. The solution to~\eqref{eq:influential} depends highly
on the influence function. To solve the problem in~\eqref{eq:influential}, the 
weight
function $\omega(\nu)=\frac{\psi(\nu)}{\nu}$ is defined, thus:

\begin{equation}
\label{eq:weight}
\sum_{j}\omega(\nu_j)\nu_j\frac{\partial \nu_j}{\partial\mathcal{X}_i}=0,
\end{equation}

\eqref{eq:weight} can be solved using the Iterative Reweighted Least Squares (IRLS)
method~\cite{zhang1997}:

\begin{equation}
\label{eq:irls}
\argmin_{\mathcal{X}}{
\sum_{j}
\omega_{k-1}(\nu_j)\nu^2_j}
\end{equation}
where, subscript $k$ indicates iteration number and $\omega_{k-1}$ is computed based on the
optimal state from the previous iteration ($k-1$).

\subsection{Least Squares with Adaptive Robust Cost Function}
\label{adaptive-cost}


Both the SC technique and the M-estimators aim to mitigate the effect of outliers 
by reducing their weight in the optimisation. The former
has the advantage of dynamically adjust the loop closures contribution by means of switch
variables which are jointly estimated in the optimisation. On the other hand, M-estimators
do not introduce any extra variables; however their parameters need to be set beforehand
and then fixed during the optimisation, regardless of the number of outliers.

Our approach, AEROS, leverages the advantages
of the M-estimators for outlier rejection, to avoid adding a switch variable to the pose graph for
every loop-closure,
while still being able to adapt its behaviour depending on the problem.~\figref{fig:sc-ad}
illustrates the difference between AEROS and the SC approach.

To achieve this behaviour, we utilise the general adaptive robust loss presented
by Barron~\cite{Barron_2019_CVPR}, which can represent a wide range of
the M-estimators by only changing a parameter. This is, in
principle,
similar to the approach of Chebrolu \etal~\cite{chebrolu2020adaptive}, however, we show
how to obtain a closed-form expression that allows us to solve the problem in a 
joint optimisation.

\subsubsection{General Adaptive Kernel}
\label{barron-kernel}

The first key part of our approach is the adaptive cost function presented by 
Barron~\cite{Barron_2019_CVPR}. It can be adjusted to model the behaviour of a large family of
robust
functions simply by varying a continuous hyper-parameter called the \textit{shape parameter} $\alpha \in \rm
I\mathbb{R}$:

\begin{equation}
\label{eq:ad-kernel}
\rho(\nu,\alpha,c)=\frac{\abs{\alpha-2}}{\alpha}\left(\left(\frac{(\nu/c)^2}{\abs{\alpha-2}}+1\right)^
{ \alpha/2}-1\right),
\end{equation}
where, $\nu$ is the weighted residual between the measurement ($z$) and its 
fitted value ($y$) represented
in~\eqref{eq:map_min} and $c>0$ is a tuning parameter which determines the size of the quadratic basin
around $\nu=0$. The tuning parameter $c$ needs to be set before optimisation.

\begin{figure}[t]
\centering
\includegraphics[width=1\linewidth,trim={2cm 0cm 2cm 0cm},clip]{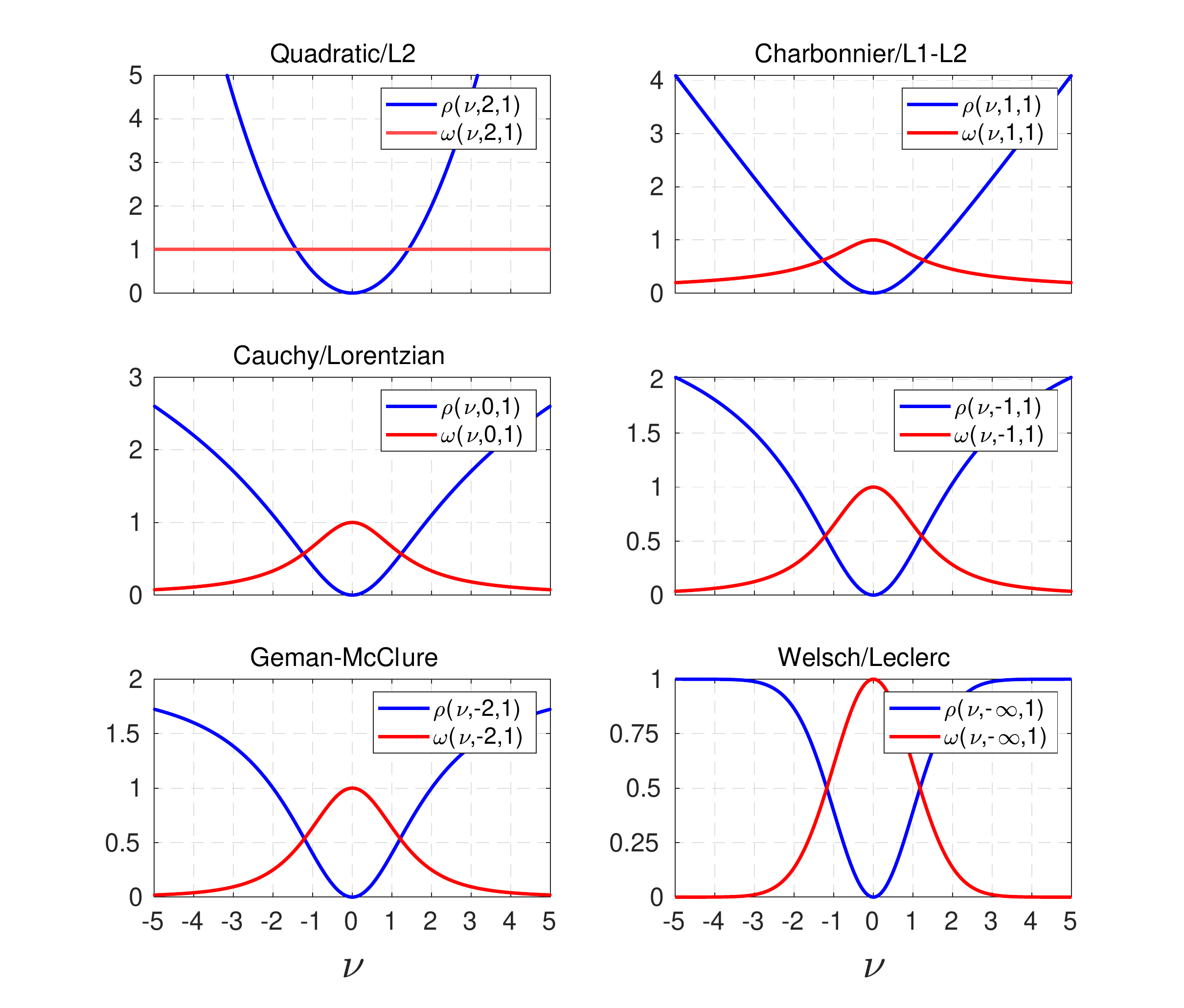}\\
\caption{\small{The adaptive loss kernel and its weight for various values of the
shape parameter $\alpha$. When $\alpha$ is 2, the adaptive loss kernel forms the standard quadratic/L2
function which is not robust to outliers. For $\alpha$ equal to 1, 0, -2 and $-\infty$}, the
adaptive loss kernel can form the robust functions known as Charbonnier/L1-L2, Cauchy/Lorentzian,
Geman-McClare and Welsh/Leclerc, respectively.}
\label{fig:kernels-weights}
\vspace{-2mm}
\end{figure}

As shown in~\figref{fig:kernels-weights}, Barron's adaptive loss kernel can represent a wide range of
well-known robust loss functions for different values of $\alpha$.
As discussed in \cite{Barron_2019_CVPR}, the adaptive kernel is undefined if $\alpha = \{2,0\}$. These
singularities need to be considered during the optimisation of the shape
parameter. However, the adaptive kernel approaches
quadratic/L2 loss function or Cauchy kernel~\cite{black1996robust} if $\alpha$ tends to 2 or 0,
respectively. If $\alpha = 1$, the adaptive kernel, $\rho(\nu,1,c)=\sqrt{(\nu/c)^2+1}-1$,
resembles L2 squared loss near $\nu=0$ and L1 absolute loss when $\nu$ is large. Due to this
behaviour, the robust kernel with $\alpha=1$ is known as an L1-L2 loss~\cite{zhang1997} or
a pseudo-Huber since it is a smooth approximation of the Huber loss function~\cite{huber1992robust}.

The adaptive loss kernel is presented as:

\begin{equation}
\label{eq:ad-kernel-with-singularities}
\rho(\nu,\alpha,c)=
\begin{cases}
0.5(\nu/c)^2 & \text{if } \alpha\rightarrow2 \\
\log\left(0.5(\nu/c)^2+1\right) & \text{if }  \alpha\rightarrow0\\
1-\exp\left(-0.5(\nu/c)^2\right) & \text{if } \alpha\rightarrow-\infty\\
\frac{\abs{\alpha-2}}{\alpha}\left(\left(\frac{(\nu/c)^2}{\abs{\alpha-2}}+1\right)^
{ \alpha/2}-1\right) & \text{else}
\end{cases}
\end{equation}

The weight $\omega(\nu,\alpha,c)$, is the derivative of $\rho$ w.r.t.
the residual $\nu$ over the residual, i.e. $\omega = \frac{1}{\nu}\frac{\partial \rho}{\partial
\nu}$, and is defined as:

\begin{equation}
\label{eq:ad-weight-with-singularities}
\omega(\nu,\alpha,c)=
\begin{cases}
1/c^2 & \text{if } \alpha\rightarrow2 \\
2/(\nu^2+2c^2) & \text{if }  \alpha\rightarrow0\\
\frac{1}{c^2}\exp\big(-0.5(\nu/c)^2\big) & \text{if } \alpha\rightarrow-\infty\\
\frac{1}{c^2}\bigg(\frac{(\nu/c)^2}{\abs{\alpha-2}}+1\bigg)^
{ (\alpha/2-1)} & \text{else}
\end{cases}
\end{equation}

We take advantage of ~\eqref{eq:ad-weight-with-singularities} to reweight the 
loop-closure
constraints and to solve the minimisation using the IRLS approach.

\subsubsection{Black-Rangarajan Duality} 
\label{black-rang}
Using the general adaptive kernel,~\eqref{eq:ad-kernel}, directly in the optimisation problem 
effectively treats the entire loop closures as outliers by estimating an $\alpha$ that
downweights all the residuals. To avoid this trivial solution,
Barron introduced a partition function, $Z(\alpha)$, as a penalty. However this makes it
impractical to implement the optimisation (please
refer to \cite{Barron_2019_CVPR} for further details).
Instead, Chebrolu \etal~\cite{chebrolu2020adaptive} presented an approximation
of $Z(\alpha)$ by means
of a truncated partition function which was pre-computed as a look-up table, and used in an
Expectation-Maximisation
fashion.

In this work, we instead
reformulate the problem using the Black-Rangarajan
duality~\cite{black1996unification} to estimate the shape and pose parameters in a joint
optimisation. Black and Rangarajan introduced a new
variable $\omega \in
[0,1]$ such that the solution of~\eqref{eq:m_estimators} remains unchanged
at the minimum. Hence a new objective function is defined as follows:

\begin{equation}
 \label{eq:black-rang}
 \rho(\nu_j)=\omega_j\nu_j^2+\Psi(\omega_j)
\end{equation}

where, $\Psi(\omega)$ is a penalty term, called an outlier process, whose expression depends on the
$\rho$-function. \eqref{eq:m_estimators} can be redefined as follows:

\begin{equation}
\label{eq:min-outlier-process}
 \argmin_{\mathcal{X},\omega_j\in[0,1]} \sum_{j}\omega_j\nu_j^2+\Psi(\omega_j)
\end{equation}

We aim to achieve the same form as in~\eqref{eq:m_estimators}
and~\eqref{eq:min-outlier-process}, thus we use $\omega = \frac{\rho'(\nu)}{2\nu}$ which comes from
the
differentiation of~\eqref{eq:m_estimators} and~\eqref{eq:min-outlier-process} with respect to $\nu$.

To determine $\Psi(\omega)$, a function $\phi(z)=\rho(\sqrt{z})$ satisfying
$\lim_{z\rightarrow0} \phi'(z)=1$, $\lim_{z\rightarrow\infty} \phi'(z)=0$ and $\phi''(z)<0$ is
defined (see~\cite{black1996unification} for more details).

By differentiating~\eqref{eq:min-outlier-process} with respect to $\omega$ and replacing $\omega$
with $\frac{\rho'(\nu)}{2\nu}$, we achieve:

\begin{equation}
\label{eq:si_derivative}
\begin{split}
 &\nu^2+\Psi'(\omega)=0\\
 &\nu^2 = -\Psi'\bigg(\frac{\rho'(\nu)}{2\nu}\bigg)
 \end{split}
\end{equation}

Now, by exploiting the function $\phi$ and integrating~\eqref{eq:si_derivative}, the outlier
process $\Psi(w)$ is obtained:

\begin{equation}
\label{eq:outlier_process}
 \Psi(\omega)=\phi\big((\phi'(\omega))^{-1}\big)-\omega\big((\phi'(\omega))^{-1}\big)
\end{equation}

\begin{figure}[t]
\centering
 \begin{minipage}{0.50\linewidth}
 \centering
   \includegraphics[width=1.05\linewidth,trim={0cm 0cm 0.5cm 0cm},clip]{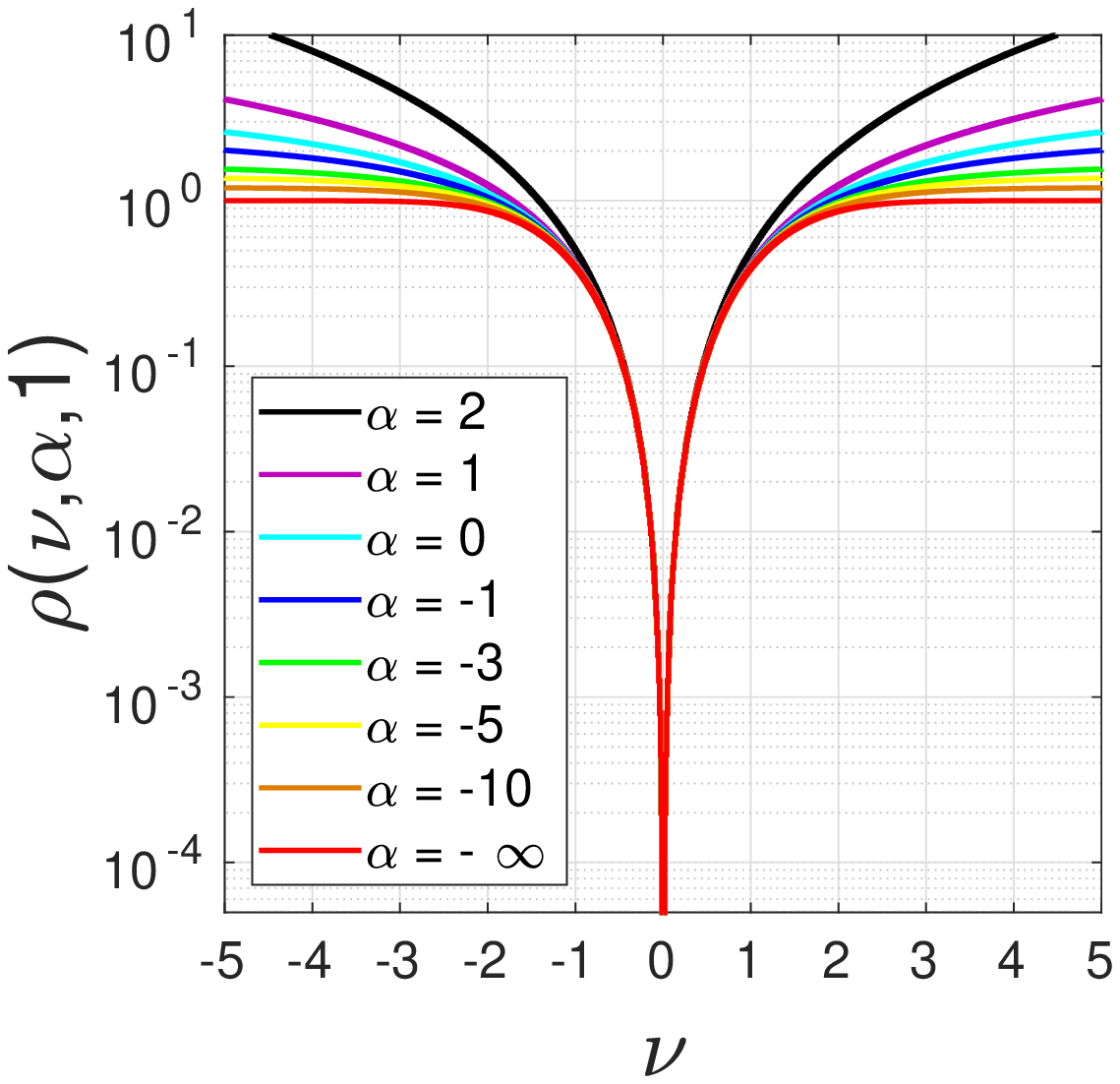}\
  \end{minipage}%
  \begin{minipage}{0.50\linewidth}
  \centering
   \includegraphics[width=1.05\linewidth,trim={0cm 0cm 0.5cm 0cm},clip]{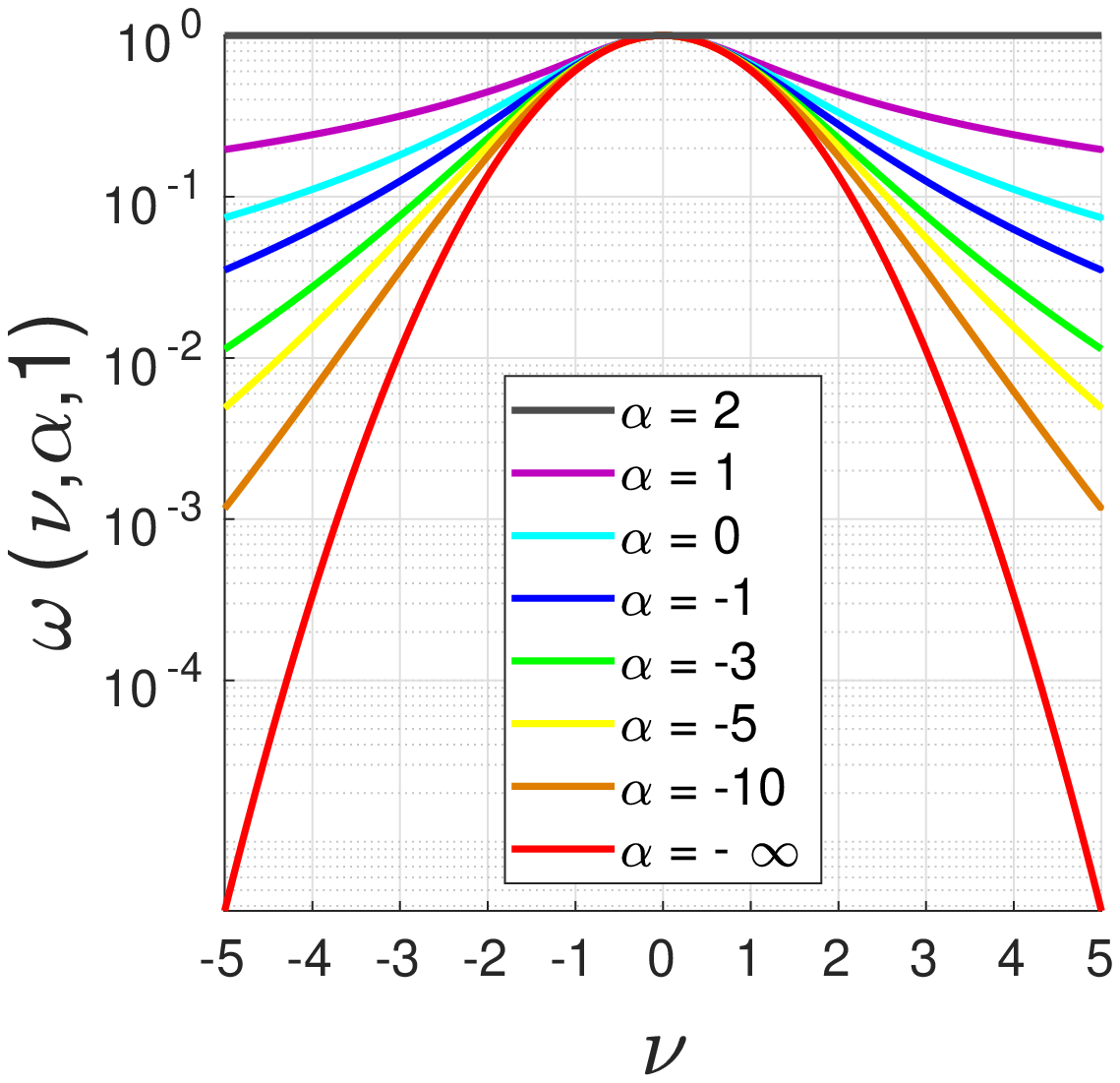}
  \end{minipage}
     \caption{\small{The logarithmic representation of the adaptive loss kernel and its weight
function for a multiple values of the shape parameter $\alpha$. It is clearly visible that the
difference between alphas smaller than -10 is negligible.}}
\label{fig:kernels}
\end{figure}
Using~\eqref{eq:outlier_process}, the outlier process corresponding to Barron's general cost
function can be derived as follows:

\begin{equation}
\label{eq:ad_outlier}
\Psi(\omega,\alpha)=
\begin{cases}
0 & \text{if } \alpha\rightarrow2 \\
-\log(\omega)+\omega-1 & \text{if }  \alpha\rightarrow0\\
\omega\log(\omega)-\omega+1 & \text{if } \alpha\rightarrow-\infty\\
\frac{\abs{\alpha-2}}{\alpha}\bigg((1-\frac{\alpha}{2})\omega^{\frac{\alpha}{\alpha-2}}+\frac{\alpha
\omega} {2}-1 \bigg) & \alpha<2
\end{cases}
\end{equation}

As seen in~\eqref{eq:ad_outlier}, the outlier process is zero for $\alpha=2$ or not defined for
$\alpha>2$ since there are no robust cost functions with these shape parameters, hence no outlier
rejection is processed.

The outlier process in \eqref{eq:ad_outlier} is a tractable, closed-form 
expression. Then, our formulation for the joint estimation of the latent 
variable $\alpha$ 
and the parameters
$\mathcal{X}$ with respect to the odometry factors $i$, loop-closure factors 
and outlier processes $j$ can then be written as:
\begin{equation}
\label{ad_algorithm}
\argmin_{\mathcal{X},\alpha}{\sum_{i}\nu_i^2 +
\sum_{j}\left(\omega_j\nu_j^2+\Psi(\omega_j,\alpha)\right)}
\end{equation}

In contrast to the SC formulation (\eqref{eq:map_min_sc}), in the optimisation 
defined in \eqref{ad_algorithm} all of the weights $\omega_j$ 
are a function of the individual parameters $\alpha$. The outlier 
process term $\Psi(\omega_j,\alpha)$ is not a prior factor but acts as a 
penalisation for $\alpha$ that precludes the optimiser from suppressing all the 
loop closures.

\begin{figure}[t]
\centering
 \begin{minipage}{1\linewidth}
 \centering
   \includegraphics[width=1\linewidth,trim={0cm 0cm 0cm 0.5cm},clip]{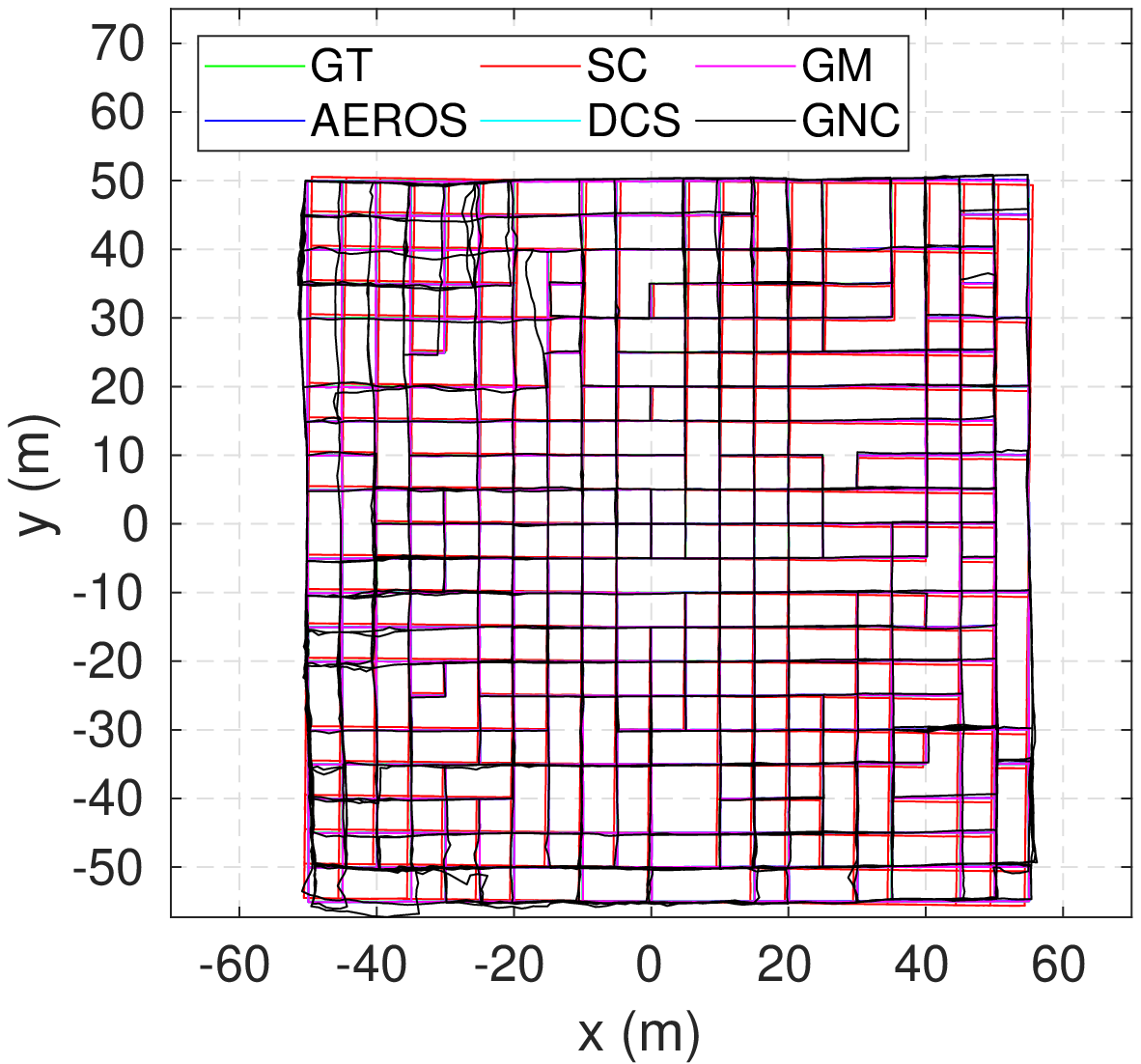}\
  \end{minipage}\\
  \begin{minipage}{1\linewidth}
  \centering
   \includegraphics[width=1\linewidth,trim={0cm 0cm 0cm 0cm},clip]{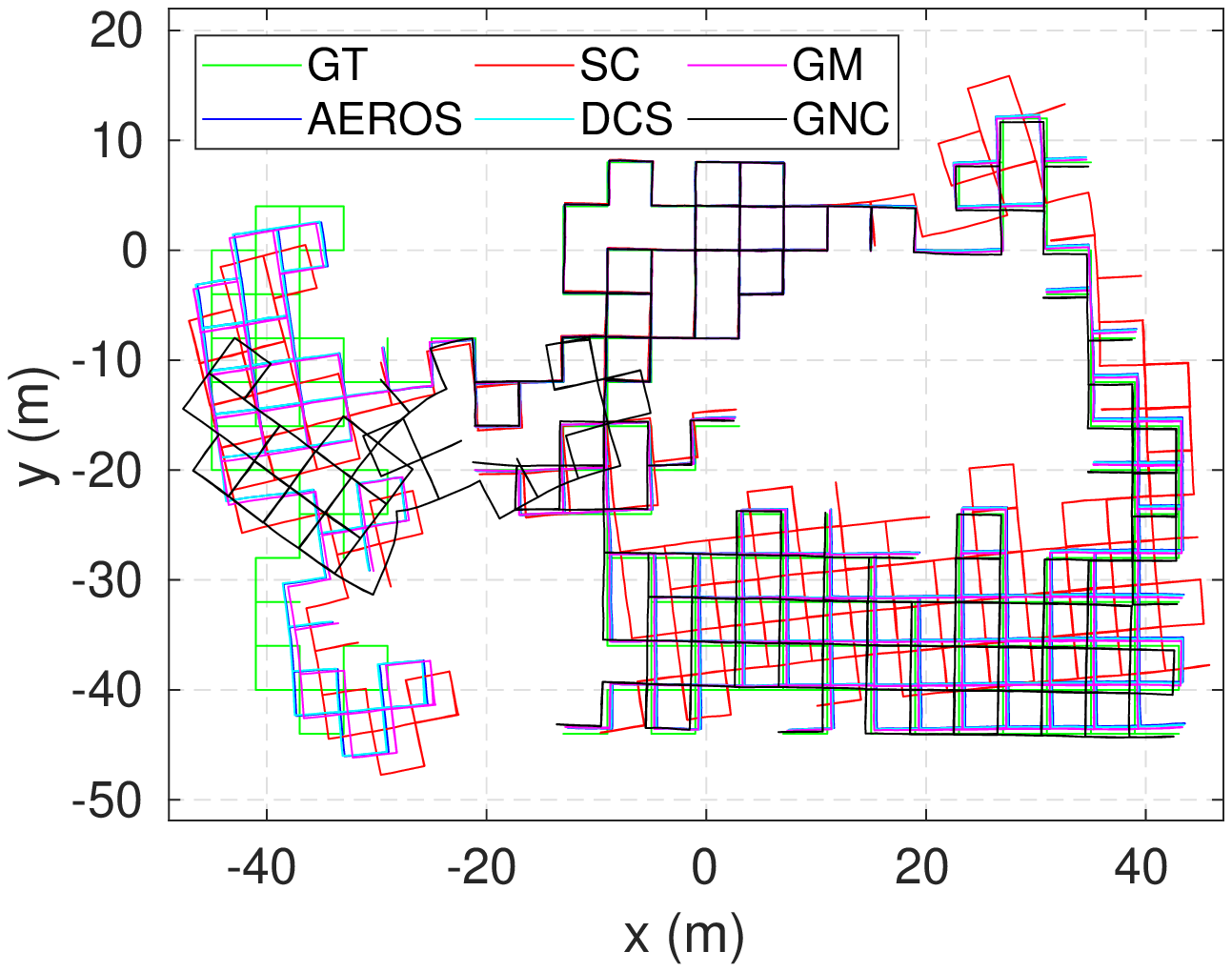}
  \end{minipage}
     \caption{\small{Estimated trajectories for the AEROS algorithm compared to
the SC, GM, DCS and GNC solutions as well as the ground truth for the City10000 (top) and
Manhattan3500
(bottom) each with $10\%$ outlier ratio.}}
\label{fig:2d-figs}
\end{figure}

While \eqref{ad_algorithm} now has a tractable expression, it has the 
disadvantage that the outlier process is not quadratic and thus cannot be used with 
standard least squares solvers. Fortunately, Rosen \etal showed that 
non-Gaussian factors can still be used if the residuals are always positive and 
are rewritten as square roots~\cite{rosen2013robust}, which is true in our case. Hence, 
the equivalent nonlinear least squares problem becomes:


\begin{equation}
\label{ad_factors}
 \argmin_{\mathcal{X},\alpha}\sum_{i}\nu_i^2
 +\sum_{j}\left(\omega_j\nu_j^2+\left(\sqrt{\Psi(\omega_j,\alpha)}\right)^2\right)
\end{equation}

We note that the generalised cost function, as well as its derived weight
(in~\figref{fig:kernels}), shows negligible change for very small values of 
$\alpha$, e.g.
smaller than -10 in our implementation. Thus, we bound the range of $\alpha$ from 2 to -10 with an
appropriate variance, e.g. 20, covering this range.

\begin{figure}[t!]
\centering
 \begin{minipage}{1\linewidth}
 \centering
   \includegraphics[width=1\linewidth,trim={0cm 0.0cm 0cm
0cm},clip]{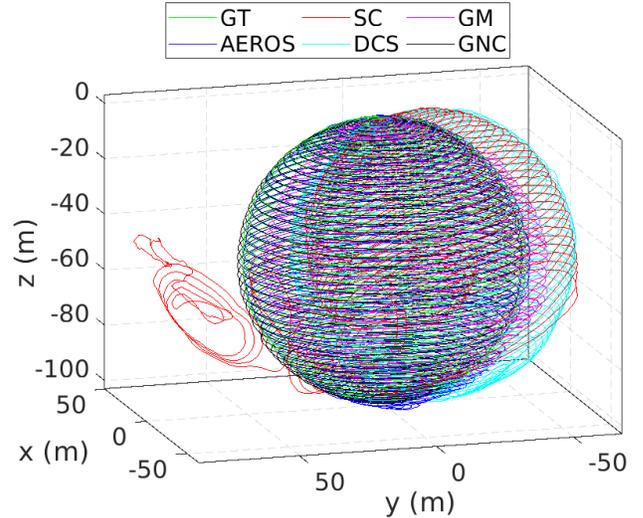}\
  \end{minipage}\\
     \caption{\small{Estimated trajectories
compared to the ground truth (green) for the
     Sphere3500 dataset with $40\%$ outlier ratio. As can be seen, the SC approach was unable to
reject some of the outliers resulting in global
inconsistency. At this Monte-Carlo run, AEROS and GNC outperform the other approaches.}}
\label{fig:3D-sphere}
\end{figure}

\section{Experiments}
\label{sec:experiments}

This section details the experiments and evaluations used to compare AEROS
to the state-of-the-art methods for outlier rejection, namely SC, DCS, 
GNC and the Geman-McClure (GM) robust loss. We tested the methods 
using widely-used public simulated and real datasets, as well as with 
real data of our own collection. All these datasets have ground truth available 
and are summarised in \tabref{table:datasets}. 

Since we are interested in real-world applications, we implemented an 
incremental SLAM system and outlier rejection methods using the iSAM2 
algorithm as a back-end. We built the system on top of the GTSAM 4.0
library\footnote{https://github.com/borglab/gtsam} except the GNC method, for
which we used the implementation available in MATLAB R2021a. All approaches were tested on a
standard PC with 4GHz Intel Xeon 4 core processor and 16 GB RAM. Our
implementation will be made open source upon publication of this paper.

Regarding the methodology, we ran 10 Monte Carlo simulations for each method on 
each dataset at different outlier ratios. We corrupted each dataset by adding 
incorrect loop closures on top of the inliers already available, from 10\% to 
50\% (see \tabref{table:datasets}). In this way we aimed to demonstrate the 
robustness of each method when new incorrect information was added but without 
removing inliers, which is more likely to occur in real applications.

For the evaluation we used the Absolute Translation 
Error (ATE) measuring directly the difference between the ground truth and the 
estimated trajectory. Additionally, to demonstrate the extent to which
the negative impact of outliers are attenuated, we used the Cumulative 
Distribution Function (CDF) to compute the
cumulative distribution of errors of each Monte Carlo run.

Lastly, we must mention the parameters used in some methods. For DCS we set 
the threshold $\Phi = 1$ as suggested in~\cite{agarwal2013robust}. 
For GNC we used Truncated Least Squares
(GNC-TLS) with \emph{TruncationThreshold} $= chi2inv(0.99,3) = 11.35$ in 2D and 
\emph{TruncationThreshold} $=
chi2inv(0.99,6) = 16.81$ in 3D, and \emph{MaxIterations} $= 1000$. GNC also 
requires to choose a local solver, for which we used the
$g^2o$~\cite{kummerle2011g} option available in MATLAB.

\subsection{Synthetic Datasets}
\label{simulated-data}

Our first experiments considered synthetic datasets commonly used to test SLAM 
systems: Manhattan3500, City10000 and Sphere2500. We used a version of 
Manhattan3500 provided by Olson~\etal~\cite{olson2006fast} and the
City10000 and Sphere2500 Datasets released by Kaess~\etal~\cite{kaess2007isam} 
with the iSAM package. As previously mentioned, for each dataset we executed 10 
Monte Carlo runs introducing random loop closures with different ratios of outliers.

We compared the result of our approach with Switchable Constraints 
(SC)~\cite{sunderhauf2012towards}, Graduated Non-Convexity
(GNC)~\cite{yang2020graduated}, as well as Dynamic
Covariance Scaling (DCS)~\cite{agarwal2013robust} and the Geman-McClure 
M-estimator.

\begin{table}[b!]
  \centering
  \caption{\small{2D and 3D datasets used in our experiments.}}
  \resizebox{\linewidth}{!}{%
  \begin{tabular}{c|ccc}
  \hline
   \multirow{2}{*}{Dataset}& Number of& Number of & Outlier \\
    & Poses& Inliers & Ratio \\
   \hline \hline
   Manhattan3500  & \multirow{2}{*}{3500}  & \multirow{2}{*}{2099}  & \multirow{2}{*}{10\%,
20\%, 30\%, 40\% and 50\%} \\
   (2D, simulated)&&& \\
   CSAIL  & \multirow{2}{*}{1045}  & \multirow{2}{*}{128}  & \multirow{2}{*}{10\%,
20\%, 30\%, 40\% and 50\%} \\
   (2D, simulated)&&& \\
   INTEL  & \multirow{2}{*}{943}  & \multirow{2}{*}{895}  & \multirow{2}{*}{10\%,
20\%, 30\%, 40\% and 50\%} \\
   (2D, simulated)&&& \\
   City10000  & \multirow{2}{*}{10000} & \multirow{2}{*}{10688} & \multirow{2}{*}{10\%} \\
   (2D, simulated)&&& \\
   \hdashline
   Sphere2500 & \multirow{2}{*}{2500}  & \multirow{2}{*}{2450}  & \multirow{2}{*}{10\%, 20\%,
30\%, 40\% and 50\%} \\
   (3D, simulated)&&&\\
   Newer College & \multirow{2}{*}{1147}  & \multirow{2}{*}{98}    &\multirow{2}{*}{100\%}   \\
   (3D, real data) &&&\\
   \hline
  \end{tabular}
  }
  \label{table:datasets}
\end{table}

\begin{figure*}[t]
\centering
 \begin{minipage}{0.5\linewidth}
 \centering
   \includegraphics[width=1\linewidth,trim={0cm 0cm 0cm
0cm},clip]{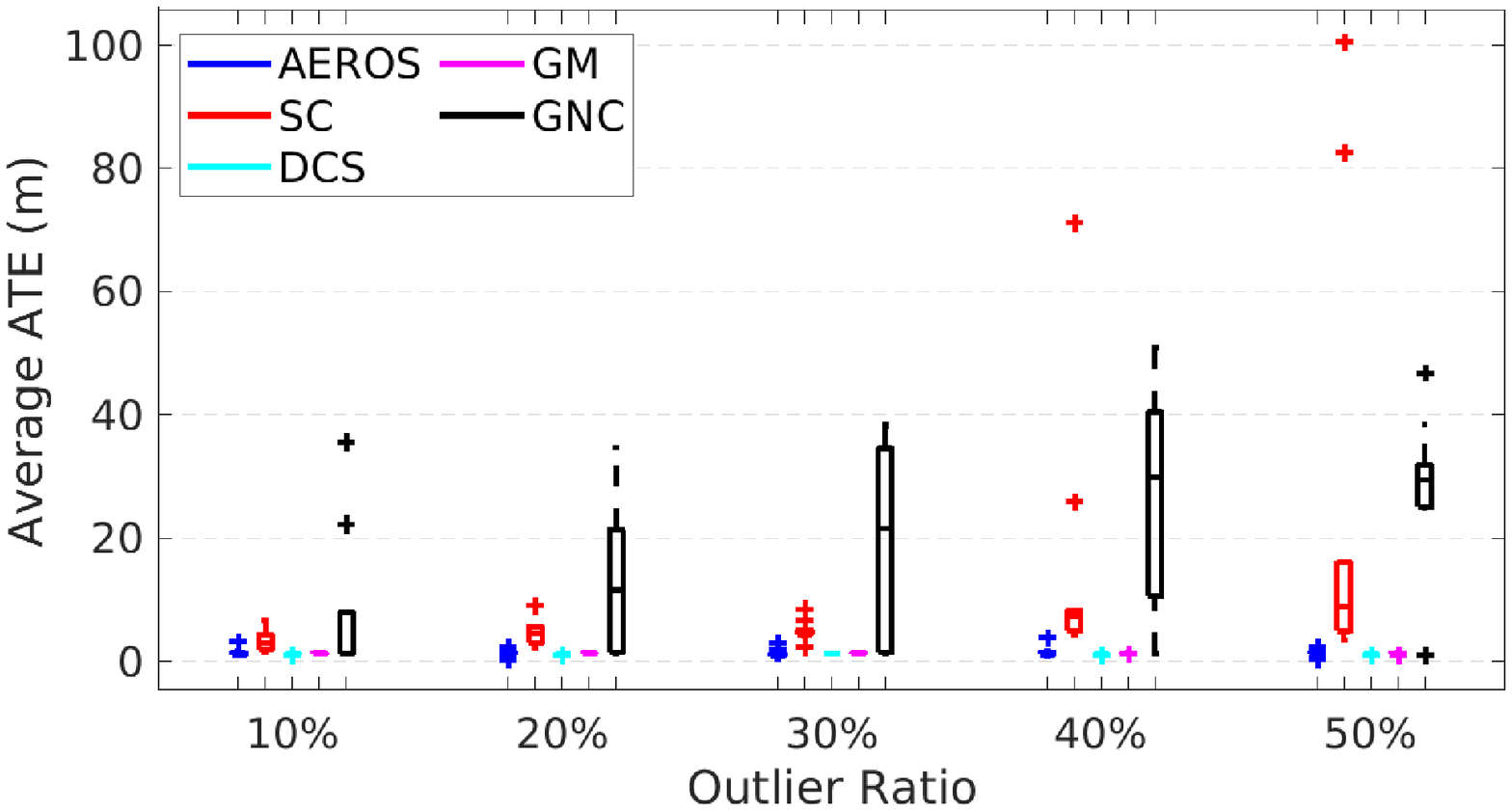}\
  \end{minipage}%
  \begin{minipage}{0.5\linewidth}
  \centering
   \includegraphics[width=1\linewidth,trim={0cm 0cm 0cm
0cm},clip]{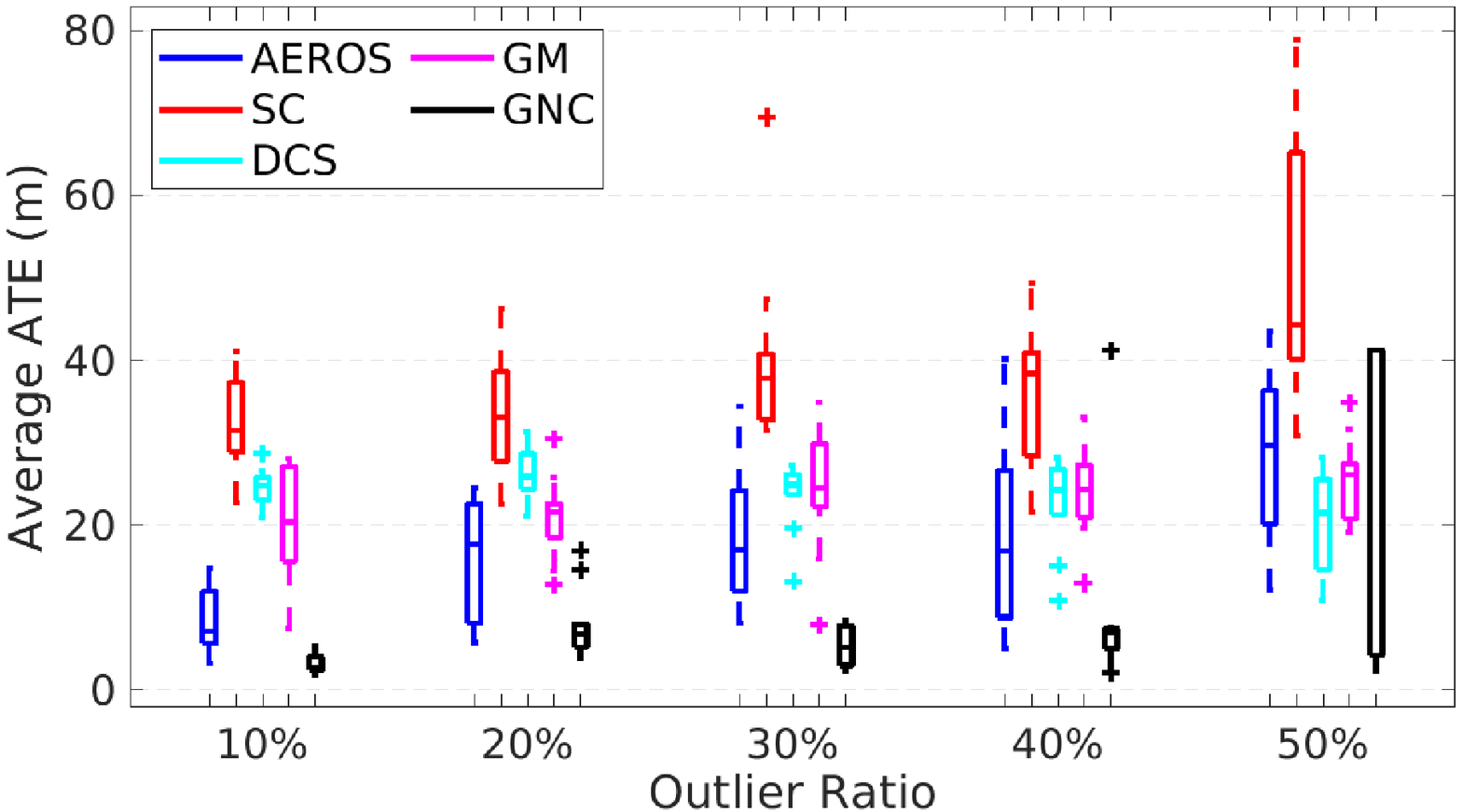}
  \end{minipage}
  \begin{minipage}{0.5\linewidth}
  \centering
   \includegraphics[width=1\linewidth,trim={0cm 0cm 0cm
0cm},clip]{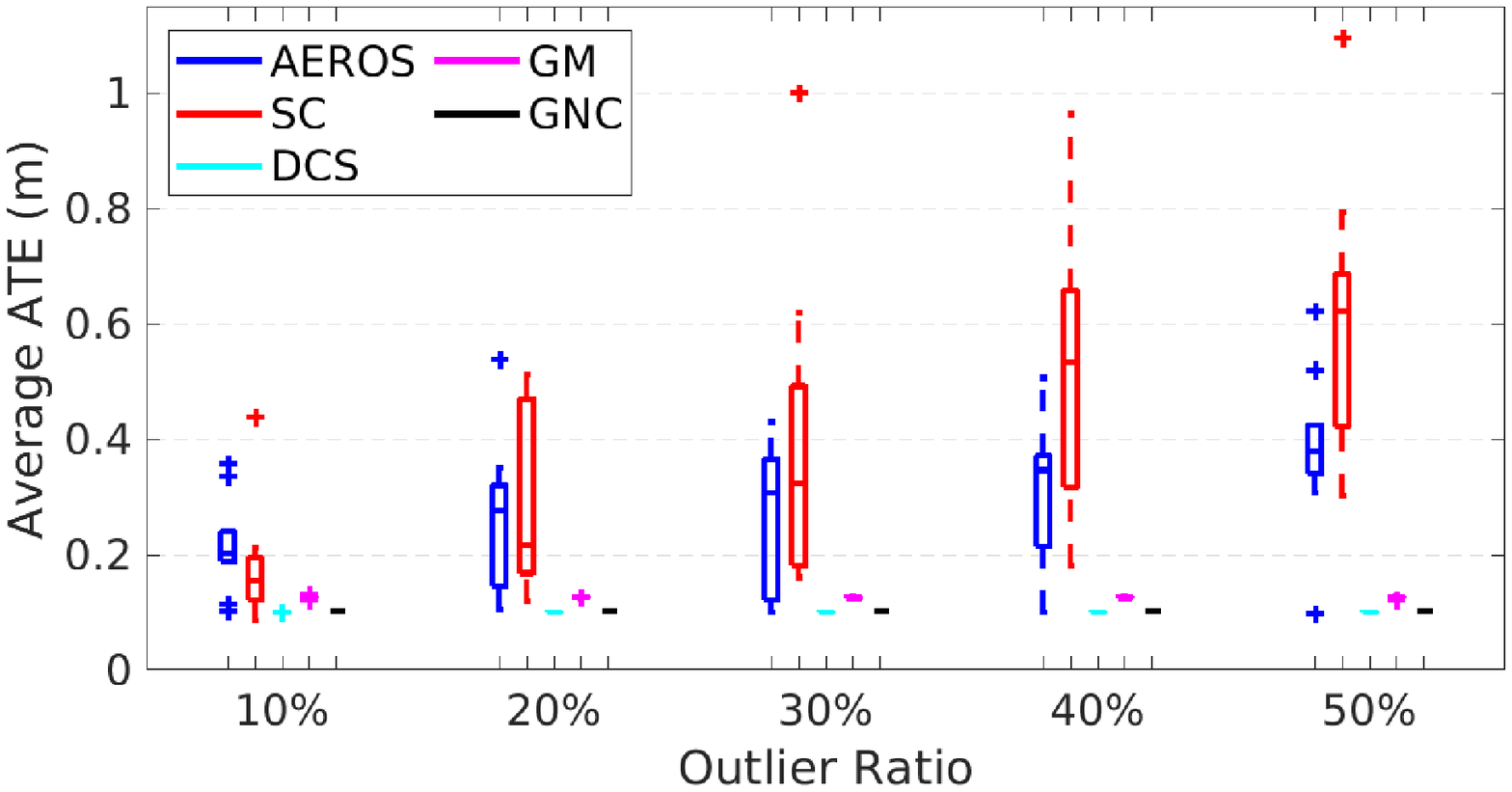}
  \end{minipage}%
  \begin{minipage}{0.5\linewidth}
  \centering
   \includegraphics[width=1\linewidth,trim={0cm 0cm 0cm
0cm},clip]{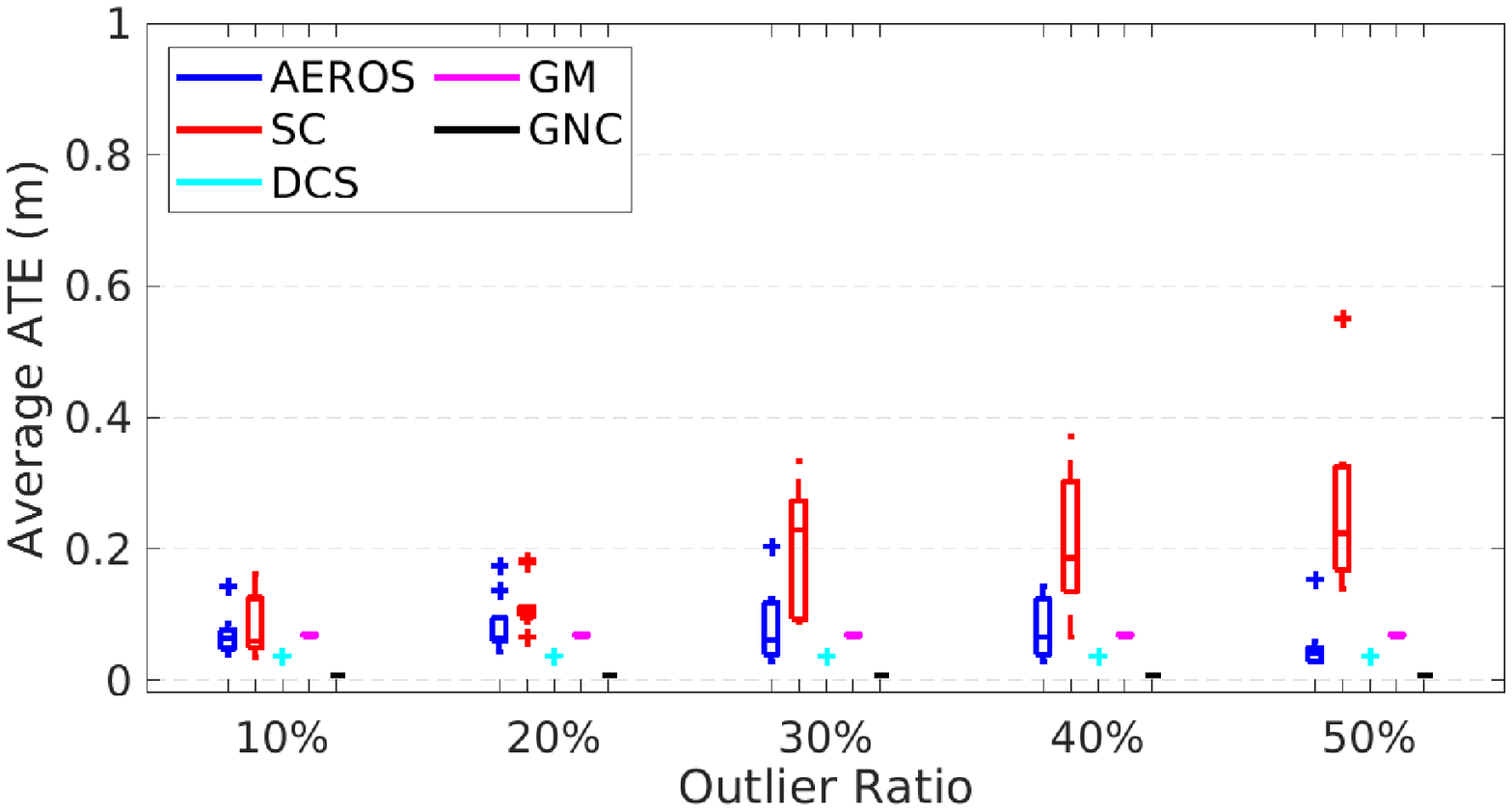}
  \end{minipage}
     \caption{\small{Performance of AEROS compared with the state-of-the-art 
     techniques for increasing outlier ratios. An average trajectory error for 
     (top-left) 2D Manhattan3500 dataset, (top-right) 3D Sphere2500, 
     (bottom-left) 2D INTEL and (bottom-right) 2D CSAIL dataset. Statistics are 
     computed over 10 Monte Carlo runs.}}
\label{fig:boxplots-10runs}
\end{figure*}


Example trajectories estimated by AEROS and the other approaches
are shown in~\figref{fig:2d-figs} and~\figref{fig:3D-sphere}. Our proposed
approach shows consistent behaviour with different ratios of outliers. This 
indicates that once the shape parameter converges, i.e. the optimum robust 
kernel is obtained, the outliers are then down-weighted and as a result, they 
have minimal impact on the optimisation.


The 2D result of running Manhattan3500
(\figref{fig:boxplots-10runs}, top-left) shows the stable and consistent 
performance of the AEROS approach.
The 3D result on Sphere2500 (\figref{fig:boxplots-10runs}, top-right) shows 
that while a higher outlier ratio affects the AEROS performance, it outperforms 
the SC algorithm and is competitive with the other approaches.

The performance of the SC approach varies depending on the number of outliers 
and as reported in~\cite{sunderhauf2012robust}, the SC method does not 
guarantee that the entire false-positive constraints can be detected. The 
result for Sphere2500 (see~\figref{fig:3D-sphere})
demonstrates that some outliers were not detected by the SC approach and 
consequently they affect the global consistency, although the map is still 
locally consistent. In addition, the combination of a robust cost
function such as Huber with the SC method to resolve global inconsistency, as 
suggested in~\cite{sunderhauf2012robust}, supports the approach we have taken 
here.

While the DCS and GM methods show stable performance when tested on Manhattan3500 and 
relatively consistent behaviour on Sphere2500, the GNC approach was 
unstable on these datasets for different outlier ratios. This is 
likely to be because the measurements have high covariance so 
setting the threshold suggested by the $\mathcal{X}^2$ distribution
causes the GNC algorithm to reject even inliers. Nonetheless, the DCS and GM 
were tested on the same datasets without being affected by this issue.

\begin{figure}[t]
 \centering
 \includegraphics[width=1.0\linewidth,trim={0.4cm 0cm 1cm
0cm},clip]{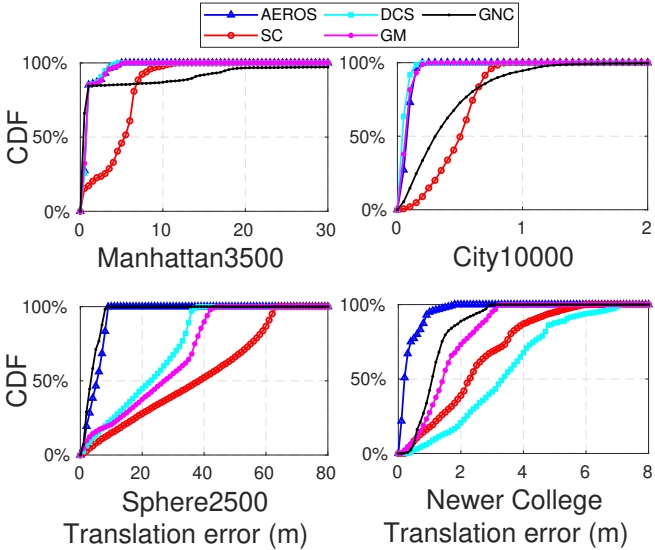}\
   \caption{\small{Cumulative distribution of errors for the experiments
described in \secref{sec:experiments}.}}
\label{fig:cdfs}
\end{figure}

So as to quantitatively measure performance of local and global consistency, we computed the cumulative distribution of the pose errors for the algorithms, which are shown in ~\figref{fig:cdfs}.
The cumulative distribution of errors for City10000 with $10\%$ outlier ratio 
indicates that all the approaches achieved local consistency. Globally, AEROS, 
DCS and GM slightly outperform the SC and GNC algorithms due to a
rotation in the maps produced by these algorithms. This could have been caused 
by an outlier that was not properly detected and down-scaled when the outlier 
ratio was low.

The cumulative distribution of
errors for Manhattan3500 also shows that our approach, along with DCS and GM, 
better maintained global consistency in the less connected regions on the left 
of the plot compared to the SC and GNC methods.

\subsection{Real 2D Data}
\label{real-2d-data}
As we are interested in the real-world application of our SLAM system, we also
tested AEROS and the other approaches on the CSAIL and
INTEL datasets described in~\cite{rosen2019se}. However, these datasets do not 
provide false-positive loop closures, so we randomly added
outliers with the same procedure as for the synthetic datasets.
\figref{fig:boxplots-10runs} (bottom) shows the results for INTEL (left) and 
CSAIL (right). Although AEROS demonstrates consistent 
performance with a range of ATE between $\sim$3 cm to $\sim$20 cm, the DCS, GM and
GNC
algorithms achieved stable ATE accuracy of $\sim$4cm, $\sim$7cm and
$\sim$1cm, respectively, over CSAIL. The range of ATE for AEROS over the
INTEL dataset is 10 cm to 60 cm, while for DCS, GM and GNC, ATE is stable at about 10 cm, 12
cm and 10 cm, respectively. This tolerance in ATE
indicates that AEROS was unable to achieve optimal performance 
due to a variance in the $\alpha$ estimate. This requires further
investigation.

\subsection{Real 3D Data}
\label{real-3d-data}
To test our algorithm in real scenarios, we used our recently published dataset 
which is described in~\cite{ramezani2020newer}. The dataset was collected using 
a handheld device consisting of a 10 Hz Ouster OS1-64 LiDAR
and a 30 Hz Intel RealSense D435i stereo camera, each of which have a built-in 
IMU. We used the long experiment (experiment \#2) from the dataset. This
experiment includes the trajectory stretching from a college quad with 
well-structured area to a garden area with dense foliage and minimal built
structure. This wide variety of environments poses a serious challenge for 
visual and LiDAR-SLAM systems.

Our LiDAR-SLAM system estimates the motion of the device using Iterative 
Closest Point (ICP)~\cite{besl1992method} at 2 Hz using visual odometry as a 
motion prior~\cite{ramezani2020online}.
To detect loop closures, we use both geometric- and appearance-based means to 
propose loop-closures which are verified by ICP registration.
The geometric method searches for loop closure candidates based on the
spatial distance, whereas, our appearance-based method employs a bag-of-words
approach~\cite{galvez2012bags} using ORB features~\cite{rublee2011orb}.

\begin{figure}[t!]
\centering
 \begin{minipage}{0.5\linewidth}
 \centering
   \includegraphics[width=1\linewidth,trim={0cm 0cm -1cm 0cm},clip]{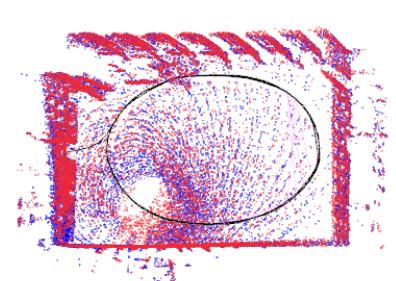}\
   \caption*{(a)}
  \end{minipage}%
  \begin{minipage}{0.5\linewidth}
  \centering
   \includegraphics[width=0.9\linewidth,trim={0cm 0cm 0cm 0cm},clip]{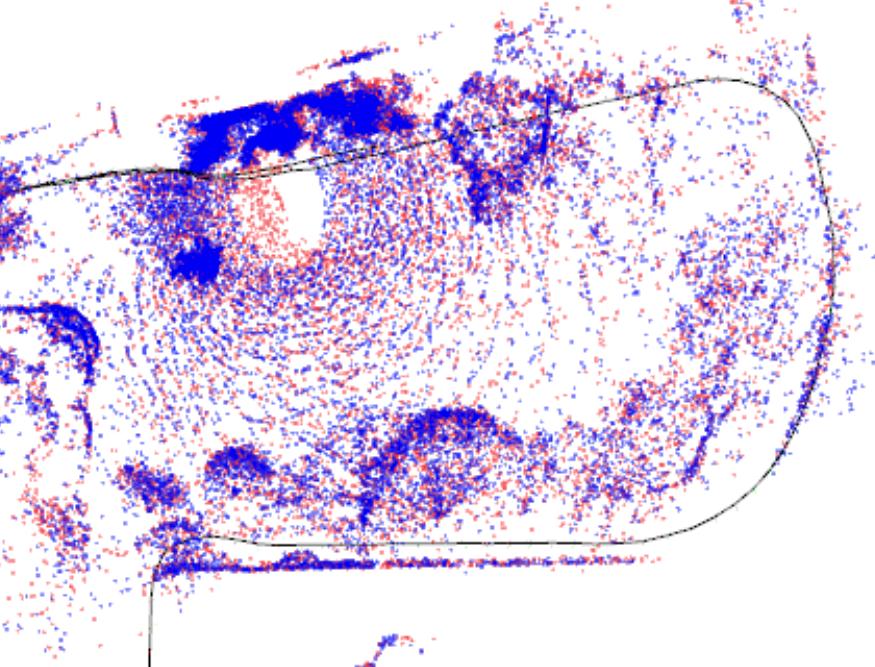}
   \caption*{(b)}
  \end{minipage}\\
  \vspace{2mm}
  \begin{minipage}{0.5\linewidth}
 \centering
   \includegraphics[width=1\linewidth,trim={0cm 0cm 0cm 0cm},clip]{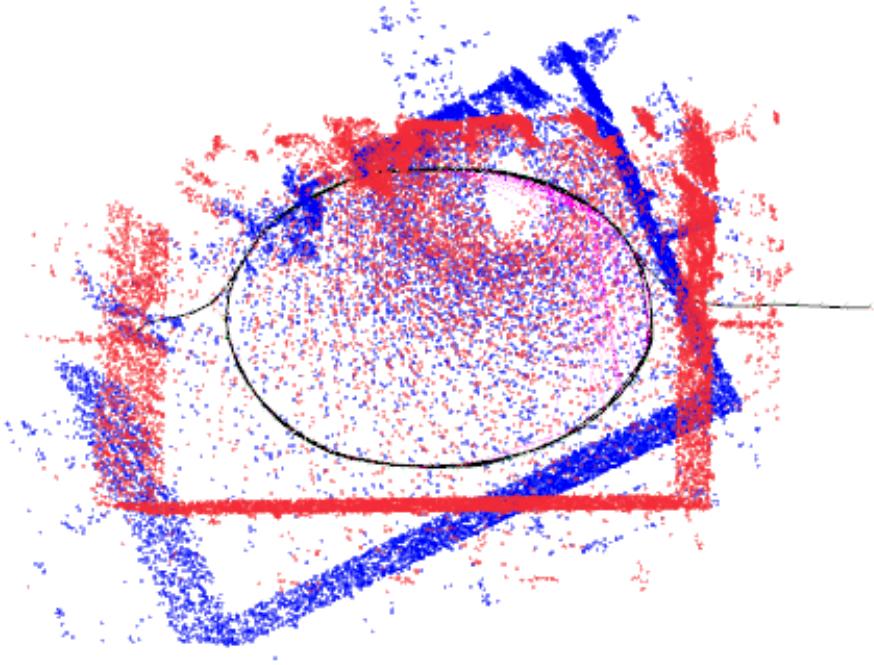}\
   \caption*{(c)}
  \end{minipage}%
  \begin{minipage}{0.5\linewidth}
  \centering
   \includegraphics[width=0.9\linewidth,trim={0cm 0cm 15cm 5cm},clip]{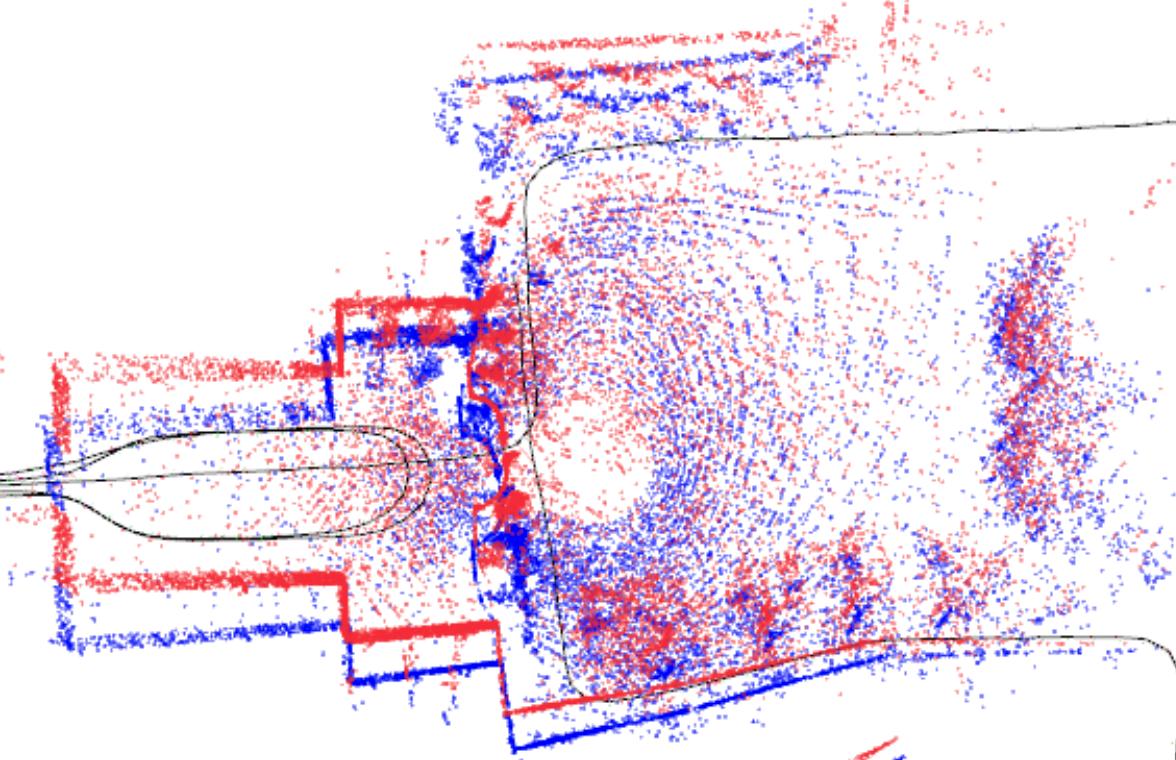}
   \caption*{(d)}
  \end{minipage}
     \caption{\small{Examples of loop closure constraints created by our (intentionally relaxed)
LiDAR-SLAM system. \textbf{Top}:
True-positive loop closures show the red and blue point clouds are properly aligned.
\textbf{Bottom}: False-positive loop closures which do not properly align the red and blue point clouds
but are plausible as many points have very low registration error.}}
\label{fig:lcs}
\end{figure}

Our LiDAR-SLAM system includes a front-end unit to filter loop closures. Once 
the candidate point clouds are registered by ICP, we analyse the point set 
(with a maximum of 20 cm point-to-plane error) and determine if there exists 
points with a diversity of normals, i.e. there is no degeneracy. We accept the 
loop candidate and it is then added to the pose graph as a 
true-positive.~\figref{fig:lcs} (a and b) shows some examples of the 
true-positive loop closures from the quad and parkland area.

As mentioned previously, a key design criterion is that loop closures are 
readily identified (for proper situational awareness) so it is
important not to set thresholds for this module very high. We want to have 
frequent true-positives yet be robust in the back-end to false-positives.

\begin{figure}[t]
 \centering
 \includegraphics[width=1.0\linewidth,trim={0cm 0cm 0cm
0cm},clip]{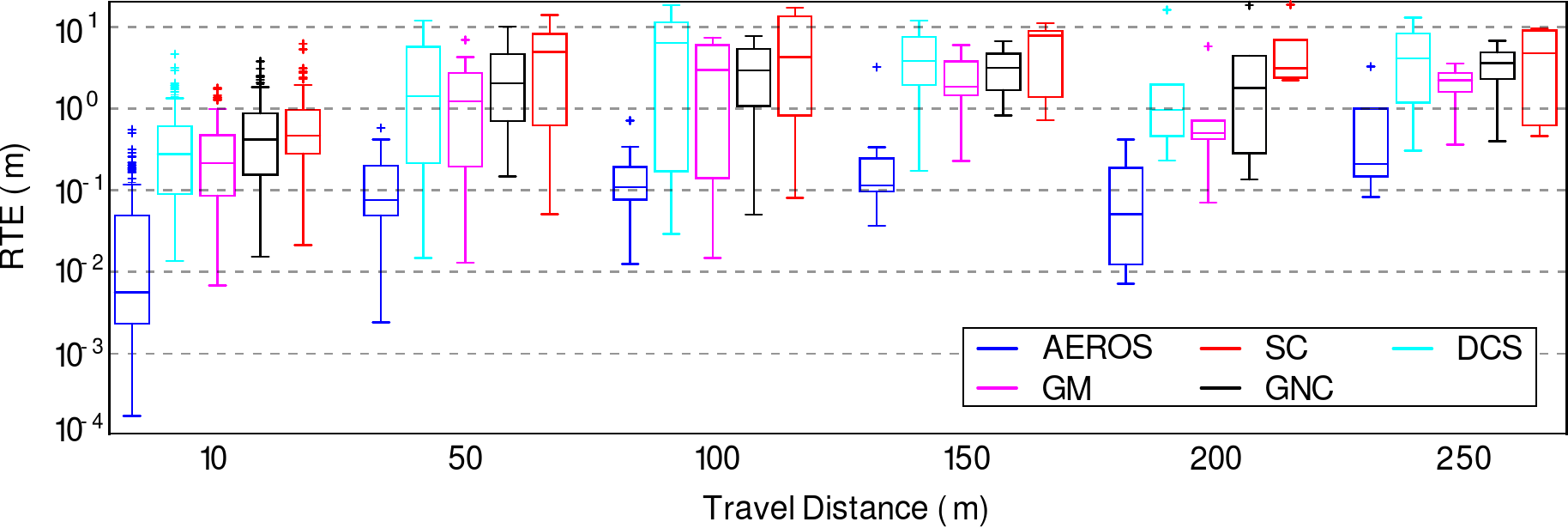}\
   \caption{\small{Comparison of AEROS versus the other approaches, DCS, SC, GM and GNC on the NCD
real data. Relative Translation Errors (RTE) are measured over various 
intervals of the trajectory of
length {10, 50, 100, 150, 200, 250} m. }}
\label{fig:rtes}
\end{figure}

In order to test the robustness of different algorithms in this dataset, we 
introduced realistic loop candidates by relaxing our appearance-based loop 
proposal mechanism. These candidates were generated using visual appearance similarity, 
which can introduce challenging loop closures to the back-end in scenes with 
repetitive structure and symmetries, such as the Quad. Having this kind of 
partly incorrect loop candidates is less likely when randomly generating 
false-positive loop closures. A few examples of these loop candidates are shown 
in~\figref{fig:lcs} (c and d). 
Note that we raised our front-end loop closure threshold such that it penalised
loop candidates and even 
candidates which may not be entirely wrong such as the one in~\figref{fig:lcs} (d). 
Loop closures that were labelled as false-positive were added as outliers to our real
dataset.

\begin{figure}[t]
\centering
 \begin{minipage}{1\linewidth}
 \centering
   \includegraphics[width=1\linewidth,trim={0cm 0cm 0cm
1cm},clip]{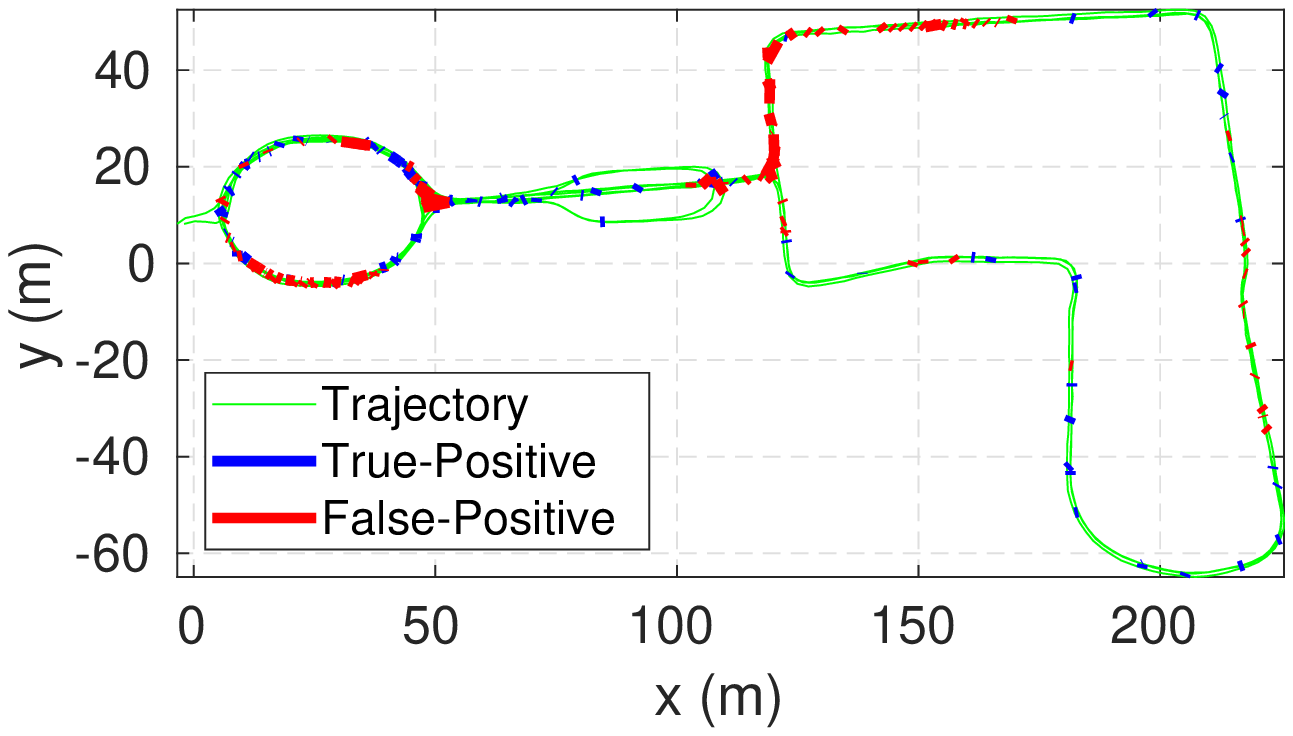}\
   \vspace{-5mm}
  \end{minipage}\\
  \begin{minipage}{1\linewidth}
  \centering
   \includegraphics[width=1\linewidth,trim={0.0cm 0cm 0cm
2cm},clip]{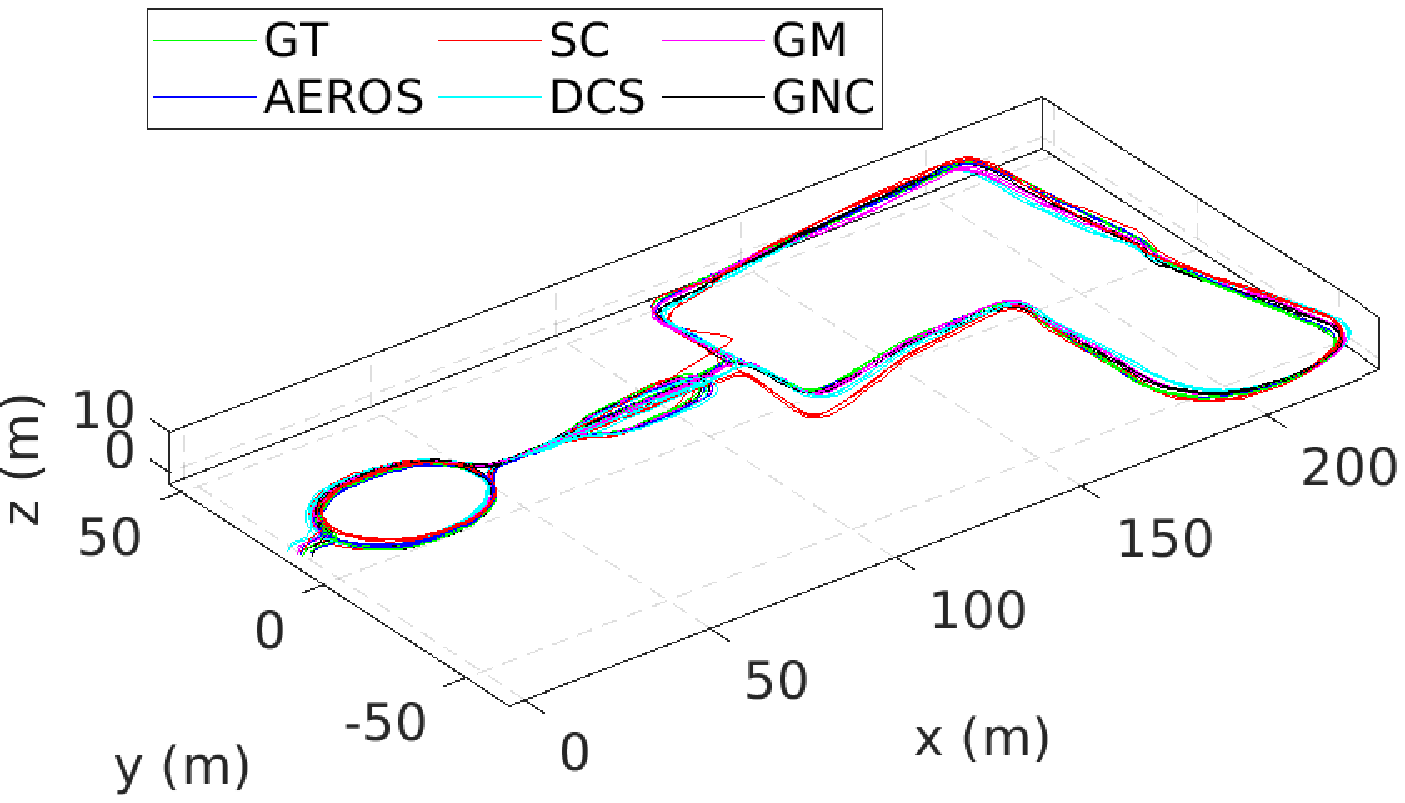}
  \end{minipage}
     \caption{\small{\textbf{Top}: Ground Truth trajectory of the Newer College dataset (the long
experiment). The blue and
red lines indicate the correct and the incorrect loop closures, respectively. Each
line connects two nodes of the pose graph --- the head and the tail of the loop
closure edge. \textbf{Bottom}: A 3D view of the trajectory estimated by the AEROS algorithm
(blue) versus
the SC solution (red), compared to the ground truth (green).}}
\label{fig:3d-new-college-long}
\end{figure}

\figref{fig:3d-new-college-long}, top, shows the ground truth trajectory of the 
experiment (green), along with the true-positive (blue) and the false-positive 
(red) loop candidates. It can be seen that the majority of rejected loop 
closures are in the parkland area (with difficult to distinguish foliage).

We ran AEROS and the other methods on the dataset and computed the ATE
(\figref{fig:cdfs}, 
bottom-right) to measure performance over the entire trajectory. In addition, we computed the
Relative Translation Error (RTE) to measure local performance at different scales
(\figref{fig:rtes}). This was done by first aligning the estimated trajectories relative to the
ground truth using Umeyama alignment~\cite{umeyama1991align} as implemented in the Python package
\emph{evo}~\cite{grupp2017evo}. As seen,
AEROS outperformed the other approaches, with
lower error than the other methods. The reason for this could be attributed to better adaptation of
the robust kernels over
the residual distribution which can then benefit from partially correct loop closures rather than
totally accepting them as inliers or fully rejecting them as outliers.
\figref{fig:3d-new-college-long}, bottom, shows example trajectories obtained
for the different methods tested.


\subsection{Computation Time}
\label{subsec:time}
\begin{figure}[t!]
\centering
 \begin{minipage}{1\linewidth}
 \centering
   \includegraphics[width=1\linewidth,trim={0.5cm 0cm 1cm
0.5cm},clip]{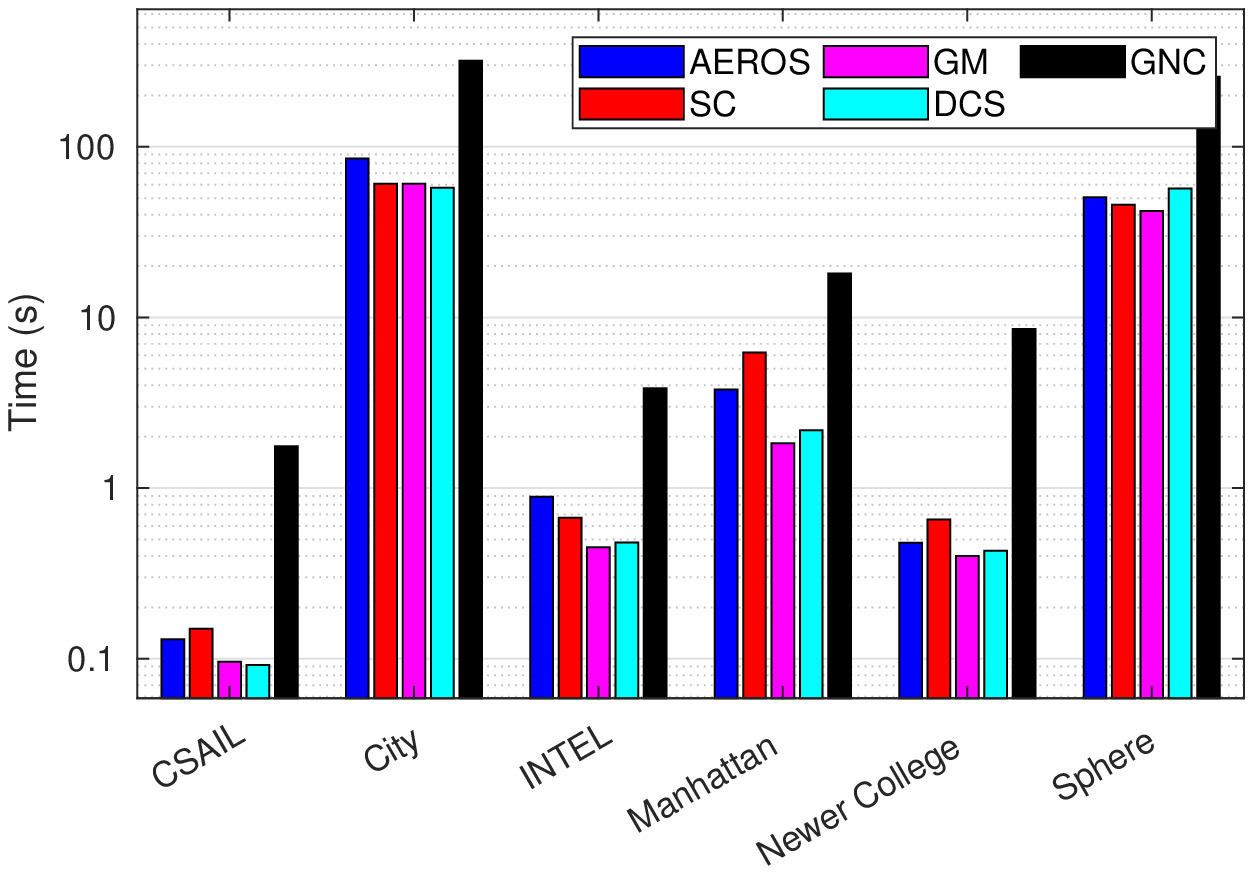}\
  \end{minipage}\\
     \caption{\small{Average computation time for the datasets used in our experiments
in~\secref{simulated-data}, \secref{real-2d-data}, and \secref{real-3d-data}.}}
\label{fig:times}
\end{figure}

\figref{fig:times} shows the average computation time over the datasets used in our experiments for
a specific outlier to noise ratio.

As noted earlier, the complexity of the AEROS
algorithm is less than the SC approach. In
AEROS the only latent parameter is the shape parameter $\alpha$ which is 
estimated along with the pose parameters, whereas SC adds a new variable for 
each loop candidate. This reduces the complexity of the underlying linear 
solver because of the reduced number of columns in the information matrix, at 
expense of the extra --though denser-- column.


Both methods require more computation time than DCS and GM, which are
10$\%$ to 30$\%$ faster because they do not introduce any extra variables in the
optimisation. Still, we consider AEROS to be competitive considering its 
adaptive properties, and it is suitable for real-time operation as observed in 
the Newer College dataset experiments. The GNC approach on the other hand 
requires more computation time compared to the other techniques.
As mentioned before, we used a MATLAB implementation of the GNC algorithm, which could affect the
run-time, while the other approaches were implemented in C++. Furthermore we used
$g^2o$ as a solver in the GNC implementation while iSAM was used for the implementation of the other approaches.


\subsection{Analysis of the Shape Parameter}
\label{subsec:alpha}
\begin{figure}[t!]
\centering
 \begin{minipage}{1\linewidth}
 \centering
   \includegraphics[width=1\linewidth,trim={0.0cm 0cm 0cm
0.0cm},clip]{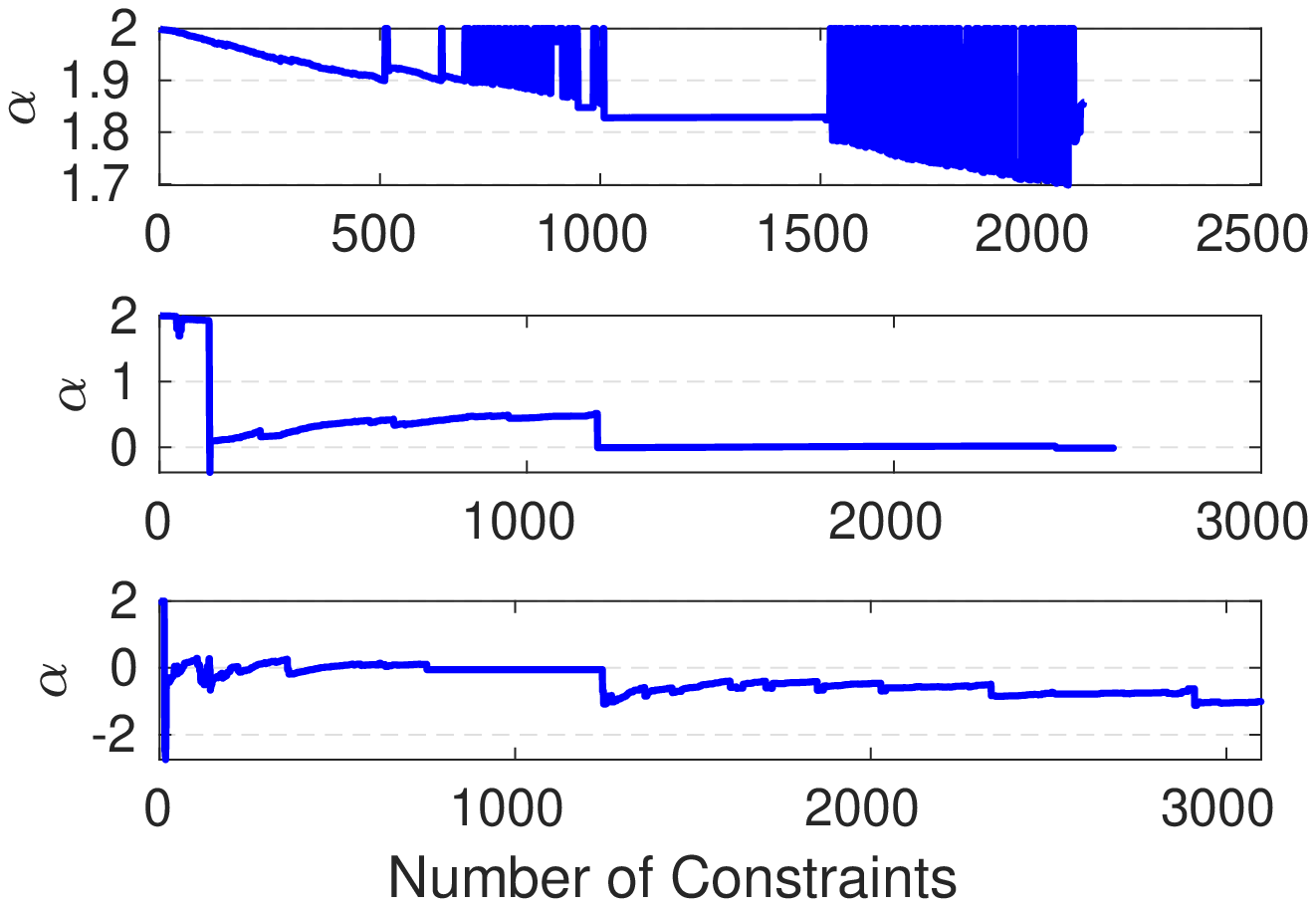}\
  \end{minipage}\\
     \caption{\small{Estimation of the shape parameter over the Manhattan3500 dataset with zero
(top), 500 (middle) and 1000 (bottom) outliers.}}
\label{fig:alpha}
\end{figure}

As previously noted, the main advantages of AEROS are due to the optimisation 
of the shape parameter $\alpha$, which allows us to cover a wide range of 
M-Estimators. To investigate the convergence of $\alpha$, we used the 
Manhattan3500 dataset with a different numbers of random outliers. As shown 
in~\figref{fig:alpha} (top), without the outliers, $\alpha$ stays relatively 
close to $2$, indicating that the adaptive cost function behaves the same as 
the standard quadratic function. By increasing the number of outliers
to $500$ (\figref{fig:alpha}-middle), the shape parameter converged to zero, 
meaning that the adaptive cost function shows a similar behaviour as the Cauchy 
fixed kernel. Lastly, we added $1000$ outliers to the dataset 
(\figref{fig:alpha}, bottom) and $\alpha$ gradually converged to $-1$, for 
which it is expected to behave similarly to the GM kernel.

\section{Conclusion and Future Work}
\label{conclusion}
In this paper, we introduced a novel approach to deal with false-positive loop 
closures in LiDAR-SLAM based on the adaptive loss function presented by 
Barron~\cite{Barron_2019_CVPR}. This adaptive loss function is able to 
represent a wide range of M-Estimators by tuning a single latent variable. 
While we formulated the pose-graph SLAM problem as a least squares minimisation, 
as is typical, we showed how to reformulate it to jointly estimate the poses and the 
latent parameter. This formulation allowed us to detect and down-weight 
false-positive loop closures without requiring an extra variable per loop-closure in the
optimisation.

We examined our approach with experiments from standard synthetic datasets, as 
well as real 2D and a large-scale 3D outdoor LiDAR dataset. Experimental
results demonstrated that our approach outperforms the Switchable 
Constraints technique \cite{sunderhauf2012towards} in both 2D
and 3D scenarios, and is competitive to other approaches such as DCS 
while being intrinsically adaptive. In future work we aim to investigate 
further the properties of the adaptive loss function in real applications, 
looking towards the development of adaptive real-time algorithms.


%

\appendices
\section{Practical Alternative to Barron's Cost Function and its Gradient and $\Psi$-Function}
As noted in~\cite{Barron_2019_CVPR}, because of the existence of singularities ($\alpha=0$ or $2$) in
the general robust kernel and instability of the kernel close to singular values, we use the following
equations in our implementation to guard against
indeterminacy when $\alpha$ converges to the singularities.

\begin{equation}
\label{eq:kernel-practical}
 \rho(\nu,\alpha,c) = \frac{p}{q}\bigg(\big(\frac{(\nu/c)^2}{p}+1\big)^{q/2}-1\bigg),
\end{equation}

\begin{equation}
\label{eq:weight-practical}
 \omega(\nu,\alpha,c) = \frac{1}{c^2}\bigg(\frac{(\nu/c)^2}{p}+1\bigg)^{q/2-1},
\end{equation}

Accordingly, the $\Psi$-function is modified as:

\begin{equation}
\label{eq:ad_outlier_practical}
\Psi(\omega,\alpha)=
\begin{cases}
0 & \text{if } \alpha\rightarrow2 \\
\frac{p}{q}\bigg((1-\frac{q}{2})\omega^{\frac{q}{q-2}}+\frac{q
\omega} {2}-1 \bigg) & \text{if }\alpha < 2
\end{cases}
\end{equation}
where, $p=\abs{\alpha-2}+\zeta$ and $\zeta=10^{-5}$. Depending on whether $\alpha$ is positive or
negative, $q$ is defined as:

\begin{equation*}
q =
 \begin{cases}
 \alpha+\zeta&\text{if }\alpha \geq0 \\
 \alpha-\zeta& \text{if } \alpha<0
\end{cases}
\end{equation*}

\section{Maximum Likelihood Least Squares Minimisation for Outlier Processes}
According to~\cite{rosen2013robust}, supposing that $f: \Omega\rightarrow
\mathbb{R}$ is a function with the corresponding factors:

\begin{equation}
 f(\alpha) = \prod_i f_i(\alpha),
\end{equation}

If $f_i(\alpha)>0$ and $\|f_i\|_\infty < c_i$, where $c_i$ is a constant $c_i \in \mathbb{R}$, we
can define:

\begin{equation}
\centering
 \begin{split}
 \nu_i: \Omega\rightarrow \mathbb{R} \\
 \nu_i(\alpha) =\sqrt{\ln c_i - \ln f_i(\alpha)}
 \end{split}
\end{equation}

Then the Maximum Likelihood solution can be estimated using Least Squares Minimisation:

\begin{equation}
 \argmax_{\alpha\in \Omega} f(\alpha) =\argmin_{\alpha\in \Omega} \sum_i \nu_i^2(\alpha)
\end{equation}

\begin{figure}[t]
\centering
 \begin{minipage}{0.5\linewidth}
 \centering
   \includegraphics[width=1\linewidth,trim={0cm 0cm 1cm 0cm},clip]{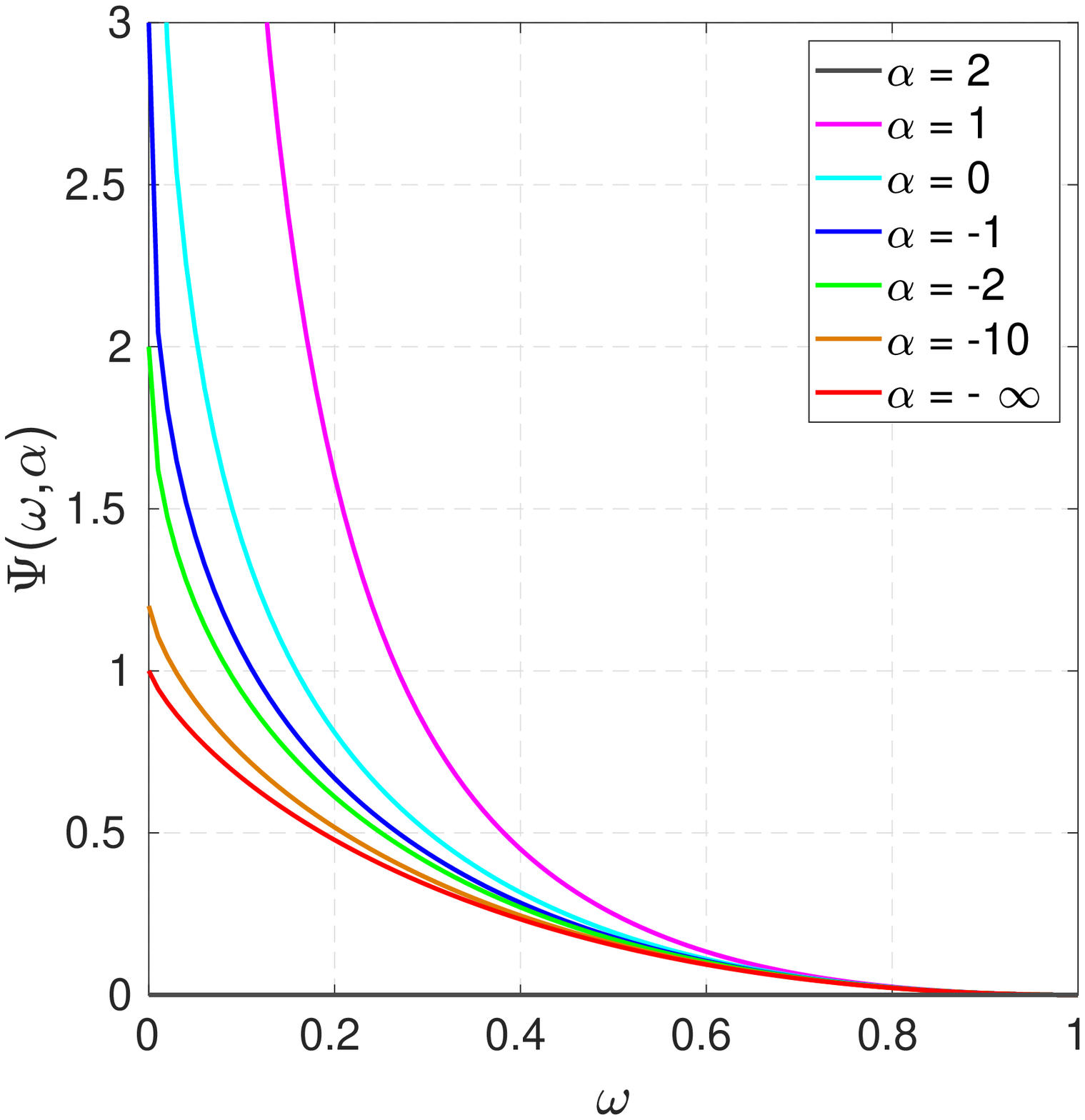}\
  \end{minipage}%
  \begin{minipage}{0.5\linewidth}
  \centering
   \includegraphics[width=1\linewidth,trim={0cm 0cm 1cm 0cm},clip]{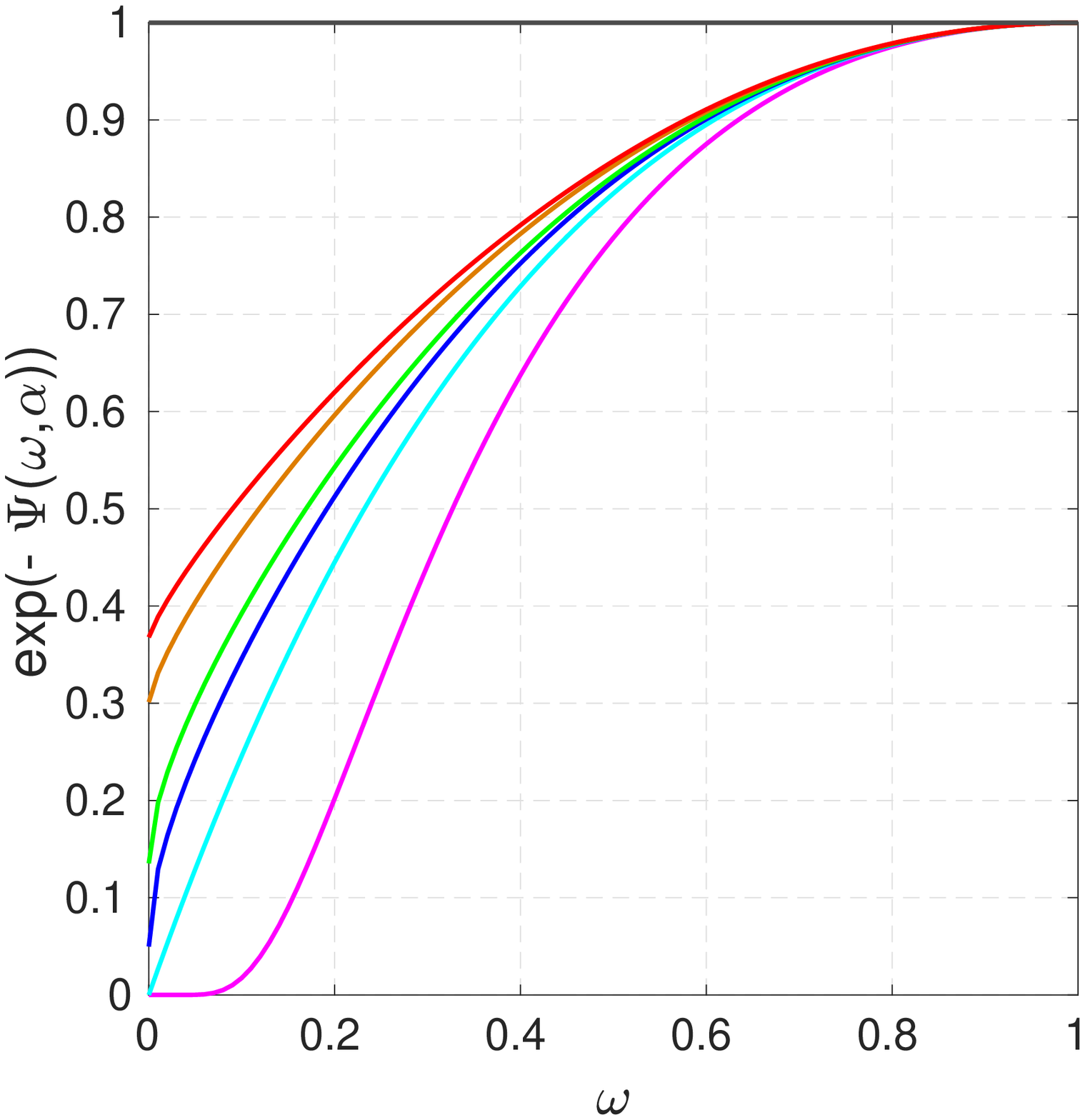}
  \end{minipage}\\
     \caption{\small{The outlier process for several values of $\alpha$ (left) and the corresponding
exponential function of the negative outlier process (right).}}
\label{fig:outlier-process}
\end{figure}

To obtain~\eqref{ad_factors}, we consider $f_i=\exp\big(-\Psi(\omega_i,\alpha)\big)$ so that
$0<f_i<1$. This is true because $\Psi$ is defined in $[0,+\infty)$
(see~\figref{fig:outlier-process}). Given $c_i = 1$, $\nu_i(\omega_i,\alpha)$ can be defined as:

\begin{equation}
\label{eq:outlier-process-LS}
 \nu_i(\omega_i,\alpha) = \sqrt{\Psi(\omega_i,\alpha)}
\end{equation}

Finally,~\eqref{eq:outlier-process-LS} can be used as the standard square residual in the least
squares minimisation expressed in~\eqref{ad_factors}.

\section*{Acknowledgment}

This research was supported by the Innovate UK-funded
ORCA Robotics Hub (EP/R026173/1). Maurice Fallon is supported by a Royal
Society University Research Fellowship. Matias Mattamala is supported by the National Agency for
Research and Development of Chile (ANID) / Scholarship Program / DOCTORADO BECAS
CHILE/2019 - 72200291.

\ifCLASSOPTIONcaptionsoff
  \newpage
\fi



%
\bibliographystyle{IEEEtran}
\bibliography{library}

%

\begin{IEEEbiography}
[{\includegraphics[width=1in,height=1.25in,trim={0cm 2cm 0cm
1cm},clip,keepaspectratio]{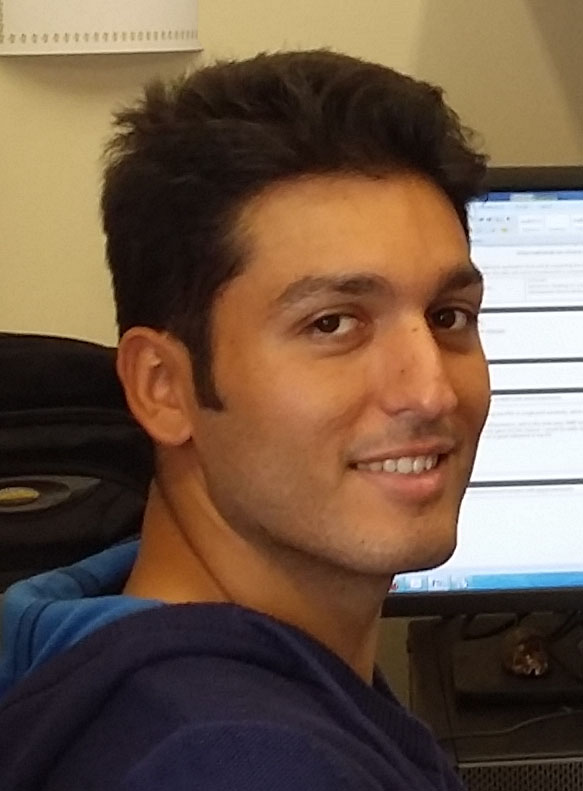}}]
{Milad Ramezani}
Milad received his PhD in the Department of Infrastructure Engineering, University of Melbourne,
Australia, 2018 in the field of Visual-Inertial Odometry.
His background is photogrammetry and he has established a system
titled ``A Vertical-Flight Surveyor'' patented by the Iranian Emblem, the Judiciary State
Organization
for registration of Deeds and Properties. He has also participated in many research projects
including aerial and satellite image processing as well as surveying and mapping.

Milad joined
the Dynamic Robots Systems Group in Oxford Robotics Institute (ORI) October, 2018, conducting his
postdoctoral research under the supervision of Dr. Maurice Fallon on 
localisation and mapping using
LiDAR, particularly for legged robots in an unknown environment. His research areas of
interest include SLAM, visual-based navigation, Kalman filter estimation, 
non-linear optimisation,
omnidirectional cameras and image processing.
\end{IEEEbiography}

\begin{IEEEbiography}
[{\includegraphics[width=1in,height=1.25in,trim={1cm 0.5cm 1cm
1cm},clip,keepaspectratio]{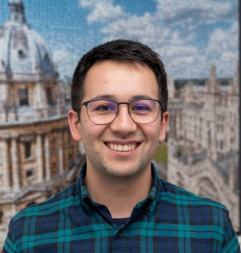}} ]
{Matias Mattamala}

Matias is a DPhil student in the Oxford Robotics Institute and a member of Hertford College since
October 2019. He is working in the Dynamic Robots Systems Group under the supervision of Dr. Maurice
Fallon. His research is focused on novel visual navigation methods for the 
ANYmal robot in challenging environments.

He completed his B.Sc, Ingenier\'ia Civil Electrica (P.Eng), and M.Sc in 
Electrical Engineering at the Universidad de Chile. His master project was 
focused on the development of a visual-proprioceptive SLAM system for a NAO 
robot. He has taught several courses at high-school and university level on 
robotics and robot perception. His research interests span from state 
estimation and computer vision to mathematical methods and geometry in robotics.

\end{IEEEbiography}


\begin{IEEEbiography}
[{\includegraphics[width=1in,height=1.25in,clip,keepaspectratio]{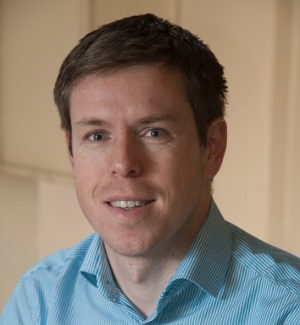}}]
{Maurice Fallon}
Dr. Fallon studied Electronic Engineering at University College Dublin. His PhD research in the
field of acoustic source tracking was carried out in the Engineering Department of the University of
Cambridge.

Immediately after his PhD he moved to MIT as a post-doc and later research scientist in the Marine
Robotics Group (2008-2012). From 2012-2015 he was the perception lead of MIT’s team in the DARPA
Robotics Challenge – a multi-year competition developing technologies for semi-autonomous humanoid
exploration and manipulation in disaster situations. The MIT DRC team competed in several phases of
the competition against world-leading international teams.

He moved to Oxford in April 2017 and took up the Royal Society
University Research Fellowship in October 2017. Currently, he leads the Dynamic Robot Systems Group
in Oxford Robotics Institute (ORI). His research is focused on probabilistic methods for
localisation and mapping. He has also made research contributions to state 
estimation for legged
robots and is interested in dynamic motion planning and control. Of particular concern is developing
methods which are robust in the most challenging situations by leveraging sensor fusion.
\end{IEEEbiography}




\end{document}